\documentclass{bmvc2k}

\title{Simpler Does It: Generating Semantic Labels with Objectness Guidance}

\addauthor{Md Amirul Islam}{cs.ryerson.ca/~amirul}{1,4}
\addauthor{Matthew Kowal}{mkowal2.github.io}{2,4}
\addauthor{Sen Jia}{github.com/SenJia}{5}
\addauthor{Konstantinos G. Derpanis}{www.eecs.yorku.ca/~kosta}{2,4,6}
\addauthor{Neil D. B. Bruce}{socs.uoguelph.ca/~brucen}{3,4}

\addinstitution{
 Ryerson University, Canada
}
\addinstitution{
 York University, Canada
}
\addinstitution{
 University of Guelph, Canada
}
\addinstitution{
 Vector Institute for AI, Canada
}
\addinstitution{
 Toronto AI Lab, LG\\
}
\addinstitution{
 Samsung AI Centre Toronto
}

\runninghead{Islam, Kowal, Jia, Derpanis, Bruce}{Pseudo-label generator}

\usepackage{times}
\usepackage{epsfig}
\usepackage{amssymb}

\usepackage{booktabs}
\usepackage{microtype}   
\usepackage{nicefrac}  
\usepackage{algpseudocode}
\usepackage{color,soul}
\usepackage{multirow,graphicx,paralist,bigdelim}
\usepackage{colortbl}
\usepackage{tabularx}
\usepackage{pgfplots}
\usepackage{enumitem}
\usepackage{amssymb}
\pgfplotsset{compat=1.3}
\definecolor{Gray}{gray}{0.85}
\definecolor{LightCyan}{rgb}{0.88,1,1}
\definecolor{antiquefuchsia}{rgb}{0.57, 0.36, 0.51}
\definecolor{bleudefrance}{rgb}{0.19, 0.55, 0.91}
\usepackage{pgfplots, pgfplotstable}
\usepackage{epstopdf}
\usepackage{epsfig}
\usepackage{pdfrender}
\usepackage{wrapfig}
\usepackage{tikz}
\usepackage{enumitem}

\usepackage{listings}
\usepackage{xcolor}

\definecolor{codegreen}{rgb}{0,0.6,0}
\definecolor{codegray}{rgb}{0.5,0.5,0.5}
\definecolor{codepurple}{rgb}{0.58,0,0.82}
\definecolor{backcolour}{rgb}{0.95,0.95,0.92}

\lstdefinestyle{mystyle}{
    backgroundcolor=\color{backcolour},   
    commentstyle=\color{codegreen},
    keywordstyle=\color{magenta},
    numberstyle=\tiny\color{codegray},
    stringstyle=\color{codepurple},
    basicstyle=\ttfamily\footnotesize,
    breakatwhitespace=false,         
    breaklines=true,                 
    captionpos=b,                    
    keepspaces=true,                 
    numbers=left,                    
    numbersep=5pt,                  
    showspaces=false,                
    showstringspaces=false,
    showtabs=false,                  
    tabsize=2
}

\usepackage{breakcites}

\definecolor{Gray}{gray}{0.90}

\definecolor{maroon}{cmyk}{0,0.87,0.68,0.32}
\newcommand{\RN}[1]{%
	\textup{\uppercase\expandafter{\romannumeral#1}}%
}

\newcommand{\blue}[1]{\textcolor{black}{#1}}
\newcommand{\bluee}[1]{\textcolor{violet}{#1}}

\begin{document}

\maketitle

\begin{abstract}
Existing weakly or semi-supervised semantic segmentation methods utilize image or box-level supervision to generate pseudo-labels for weakly labeled images. \blue{However, due to the lack of strong supervision, the generated pseudo-labels are often noisy near the object boundaries, which severely impacts the network's ability to learn strong representations}. To address this problem, we present a novel framework that generates pseudo-labels for training images, which are then used to train a segmentation model. To generate pseudo-labels, we combine information from: (i) a class agnostic `objectness’ network that learns to recognize object-like regions, and (ii) either image-level or bounding box annotations. We show the efficacy of our approach by demonstrating how the objectness network can naturally be leveraged to generate object-like regions for \textit{unseen} categories. We then propose an end-to-end multi-task learning strategy, that jointly learns to segment semantics and objectness using the generated pseudo-labels. Extensive experiments demonstrate the high quality of our generated pseudo-labels and effectiveness of the proposed framework in a variety of domains. Our approach achieves better or competitive performance compared to existing weakly-supervised and semi-supervised methods. 
\end{abstract}

\section{Introduction}\label{sec:intro}
\vspace{-0.1cm}
State-of-the-art methods for semantic segmentation~\cite{long15_cvpr,chen15_iclr,noh15_iccv,badrinarayanan15_arxiv,Islam_2017_CVPR,ghiasi2016laplacian,islam2017label,yu2015multi,ghiasi2016laplacian,zhao2017pyramid,refinenet,chen2017rethinking,chen2018deeplab,karim2020distributed,karim2019recurrent,islam2020feature,takikawa2019gated,islam2021segmix} are founded on fully convolutional networks (FCN)~\cite{long15_cvpr} to segment semantic objects in an end-to-end manner. 
A caveat of such training is that it requires supervision with an extensive amount of pixel-level annotations. Since the expense for generating semantic segmentation annotations is large, a natural solution is to address the problem of semantic segmentation with one of two common supervision settings, weakly or semi-supervised. 

In the weakly supervised semantic segmentation (WSSS) setting, labels used during training contain only partial information. Recently proposed WSSS methods utilize image-level labels~\cite{fan2020employing,chen2020weakly,chang2020weakly,fan2020learning,ahn2018learning,huang2018weakly,hou2018self,lee2019ficklenet,zeng2019joint,lee2019frame,ahn2019weakly,jiang2019integral,wei2018revisiting}, scribbles~\cite{lin2016scribblesup}, or bounding box~\cite{dai2015boxsup,khoreva2017simple,song2019box} supervision to learn semantic masks. 
\begin{figure}[t]
	\begin{center}
		\includegraphics[width=0.97\textwidth]{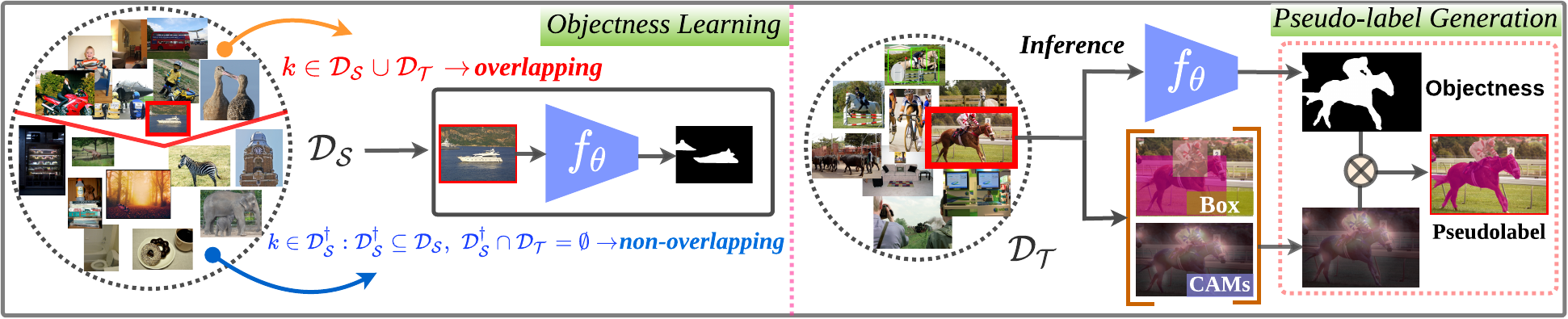}
		\vspace{-0.25cm}
		\caption{\textbf{Left:} An illustration of our process for generating high-quality semantic segmentation pseudo-labels for a target dataset, $\mathcal{D_T}$. We first train a objectness network, $f_\theta$, on a source dataset under one of two data settings, (\textit{overlapping} ($\mathcal{D_S}$) or \textit{non-overlapping} ($\mathcal{D^{\dagger}_{S}}$) categories ($k$) with $\mathcal{D_T}$), that learns to generate a class-agnostic objectness prior. \textbf{Right:} Then, we use either Class Activation Maps (CAMs)~\cite{zhou2016learning} or bounding box proposals combined with a class agnostic objectness prior to generate a pseudo-label. }
		\label{fig:pseudo}
	\end{center}
	\vspace{-0.7cm}
\end{figure}
Most of these methods rely on incorporating additional guidance to obtain the location and shape information. A common way to obtain location cues from class labels is by using \textit{Class Activation Maps} (CAMs)~\cite{zhou2016learning} as it roughly localizes semantic regions of each class. However, utilizing CAMs directly as supervision can be problematic as they roughly localize objects and cannot capture detailed object boundaries between different semantic regions. 
Recent works have addressed this issue in a variety of ways~\cite{pinheiro2015image,kwak2017weakly,ahn2018learning,ahn2019weakly}, one of the most effective being the use of object guidance via the use of class agnostic saliency~\cite{huang2018weakly,lee2019ficklenet,zeng2019joint,jiang2019integral}.
Similarly, bounding box based methods~\cite{kulharia2020box2seg,song2019box,li2018weakly,khoreva2017simple,huang2018weakly} typically rely on generating rough pseudo-labels by applying the unsupervised CRF~\cite{krahenbuhl2011efficient}, MCG~\cite{pont2016multiscale}, or GrabCut~\cite{rother2004grabcut} methods to remove irrelevant regions from the semantic region proposal in an iterative way to obtain stronger pseudo-labels at each iteration. 
However, the quality gap between the pseudo-labels and groundtruth is typically \textit{large} for the CAM-based and bounding box-based approaches. Furthermore, iterative procedures and complex pipelines can make the data generation process for these methods computationally expensive and time consuming. 

In the semi-supervised semantic segmentation (SSSS) setting, the groundtruth annotations are used but only for a fraction of the total number of training examples, e.g., 10\% of the labels~\cite{souly2017semi}. Similar to the techniques used in WSSS methods, pseudo-labels are then generated for the unlabelled data (e.g., by using additional image-level annotations~\cite{xiao2018transferable,wei2018revisiting,lee2019ficklenet,khoreva2017simple,huang2018weakly}). \blue{Recent work~\cite{hu2018learning} introduced a partially supervised training paradigm which learns to segment everything  using a portion of box and mask annotations.} However, these methods still require labour-intensive pixel-level \textit{semantic} annotations and the performance heavily depends on the \textit{quantity} of the labeled data and the \textit{quality} of the generated pseudo-labels.

In the light of the highlighted issues that arise in WSSS and SSSS methods, we propose a novel simple yet effective pipeline which transfers `objectness' knowledge to weakly labeled images for learning semantic segmentation. The intuition behind using the objectness guidance instead of widely used saliency-based approaches~\cite{oh2017exploiting,wei2018revisiting,zeng2019joint,yao2020saliency} is that groundtruth saliency masks inherently ignore objects near the border of the image due to the well-known centre bias~\cite{bruce2015computational, accik2014real}. Recent works~\cite{xiao2018transferable,wang2016objectness} also utilize the objectness prior to refine the semantic proposals.
There are two key differences between our work and~\cite{xiao2018transferable}. First is the use of a source dataset.~\cite{xiao2018transferable} obtains the objectness prior strictly from the target data distribution, which is arguably an easier problem to solve. However in our work, we strictly prohibit the use of per-pixel labels from the \textit{target} dataset and only use a \textit{source} dataset (i.e., COCOStuff) for the objectness prior. We argue that using COCOStuff as the \textit{source} data (instead of VOC like in \cite{xiao2018transferable}) will allow our objectness network to generate better pseudo-labels for a more diverse set of categories and can be generalized well to different target datasets. Second, during the segmentation network training,~\cite{xiao2018transferable} uses the semantic segmentation labels for the \textit{strong} categories (i.e., the classes used to train their objectness network), while in our settings we only use pseudo-labels when training the semantic segmentation network. 


The key component of our pipeline is the pseudo-label generation approach (see Fig.~\ref{fig:pseudo}), where we first train an objectness network on a source dataset which generates a class agnostic objectness prior. We then combine this prior with weak semantic proposals (e.g., image or box-level) to generate semantic segmentation labels for a target dataset. We further show that the objectness prior is robust enough to generalize the objectness knowledge onto categories that have \textit{never been seen} by the objectness network; when the source dataset has no class \textit{overlap} with the target dataset (i.e., the \textit{non-overlapping case}). We view the \textit{non-overlapping} setting as comparable with \textit{weak-supervision}, as the objectness model has no direct understanding of the shape of the target domain classes (unlike previous methods~\cite{oh2017exploiting,wei2018revisiting,zeng2019joint,yao2020saliency} which use overlapping groundtruth saliency annotations). In contrast, the \textit{overlapping} setting (i.e., the class agnostic source dataset contains objects found in the target dataset) is comparable (but has less supervision) to \textit{semi-supervision} as class-agnostic (i.e., binary) segmentation annotations are used. Finally, for segmentation learning, we adopt a multi-task joint-learning~\cite{cheng2017segflow,cvpr18_rank,islam2018semantics,zeng2019joint,he2021sosd} based \textbf{S}emantic \textbf{O}bjectness \textbf{Net}work (denoted as SONet), with the addition of an `objectness branch', that explicitly models the relationship between \textit{semantics} and \textit{objectness}. We summarize our main contributions as follows:
\vspace{-0.1cm}
\begin{enumerate}
  \setlength\itemsep{0.4em}
    \item We introduce a simple yet effective pseudo-label generation technique that combines a class agnostic `objectness' prior with semantic region proposals. The flexibility of our technique is demonstrated by its ability to incorporate either image \textit{or} box-level labels into the pseudo-label generation pipeline.
    \item We propose a joint learning based Semantic Objectness Network, SONet, that improves the semantic segmentation quality through objectness guidance.
    
    \item We present an extensive set of experimental results which demonstrates the effectiveness of our proposed method in both the simplicity of the pseudo-label generation process as well as the quality of the pseudo-labels. Our proposed approach achieves competitive performance compared to existing WSSS methods and outperforms SSSS methods without ever using groundtruth semantic segmentation supervision.
\end{enumerate}
\vspace{-0.2cm}

    
    
\vspace{-0.2cm}
\section{Proposed Framework}\label{sec:approach}
Our pipeline consists of two key components. First, we generate pseudo-labels for training images by combining our generated objectness prior with weak semantic proposals, which are produced from either image labels or box annotations (Sec.~\ref{sec:label_gen}). Second, we introduce our multi-task model, SONet, that jointly learns to segment both semantic categories and a binary `objectness' mask, which enforces richer boundary detail and semantic information (Sec.~\ref{sec:net}). 

\vspace{-0.1cm}
\subsection{Semantic Pseudo-Label Generation}\label{sec:label_gen}
Our pseudo-label generation process consists of two separate components. We first describe the procedure behind training the `objectness' network which is designed to obtain detailed boundary information for any object-like region. Next, we describe two different techniques for generating semantic pseudo-labels by combining the output of the objectness network with semantic region proposals, which are obtained from either image-level class labels or bounding box annotations.\\

\noindent \textbf{Training an Objectness Network.}\label{sec:obj_cls} 
Pixel objectness~\cite{xiong2018pixel} quantifies how likely a pixel belongs to an object of any class (i.e., other than “stuff” classes like background, grass, sky, sidewalks, etc.), and should be high even for objects unseen during training. We use DeepLabv3 network~\cite{chen2017rethinking}, $\mathcal{\phi_P}$, on a source dataset, $\mathcal{D_S}$, to learn an objectness prior from the `things' label. We use a weak form of the COCOStuff dataset, denoted as \textit{COCO-Binary} and consider it as the source dataset, $\mathcal{D_S}$. More specifically, we generate \textit{COCO-Binary} by \textit{removing all semantic labels from the COCOStuff dataset} so what remains is binary maps where all the \textit{things} categories are assigned to the label one, and the \textit{stuff} categories to zero. 
\blue{We then train the objectness network, $\mathcal{\phi_P}$, on the source dataset, $\mathcal{D_S}$, under two different settings which outputs a pixel-wise `objectness score' (similar to the saliency detection models). In the first setting, we include all the images from the source data, $\mathcal{D_S}$, regardless of whether the objects found in $\mathcal{D_S}$ images overlap with target data, $\mathcal{D_T}$. In the second setting, we create a subset of $\mathcal{D_S}$ by excluding those images containing any categories which overlap with $\mathcal{D_T}$ categories. We can formalize the \textit{overlapping} and \textit{non-overlapping} settings as follows:}
\begin{gather}
  \resizebox{.55\hsize}{!}{$k \in \begin{cases}
              \mathcal{D_S} \quad \text{overlapping}\\
              \mathcal{D^{\dagger}_{S}}: \mathcal{D^{\dagger}_{S}} \subseteq \mathcal{D_S}, 
              \hspace{0.2cm} \mathcal{D^{\dagger}_{S}} \cap \mathcal{D_T} = \emptyset \, \hspace{0.2cm} \text{non-overlapping,} 
   \end{cases}$}
\end{gather}
\blue{where $k$ denotes the set of object classes contained in \textit{COCO-Binary} used to train the objectness model, $\mathcal{\phi_P}$.  $\mathcal{D^{\dagger}_{S}}$ represents the \textit{non-overlapping} subset where there is no semantic category overlap between $\mathcal{D^{\dagger}_{S}}$ and $\mathcal{D_T}$. Note that the semantic annotations are solely used to generate the subset of non-overlapping data, $\mathcal{D^{\dagger}_{S}}$, and is not required for training the objectness model, $\mathcal{\phi_P}$. We believe the \textit{non-overlapping} setting is more challenging than saliency-based WSSS~\cite{oh2017exploiting,wei2018revisiting,zeng2019joint,yao2020saliency}, because those methods contain semantic overlap within the source and target data. In both settings, we train the objectness classifier using the class-agnostic segmentation groundtruth and use the binary cross entropy loss function. The main goal of the objectness classifier is to learn a strong objectness representation~\cite{islam2020shape} that contributes towards creating pseudo-labels for semantic supervision. }\\

\noindent \textbf{Class-Driven Pseudo-labels.}\label{sec:label_cam} 
CAM~\cite{zhou2016learning} is widely used as a weak source of supervision as it roughly localizes semantic object areas. Following previous works~\cite{ahn2018learning,ahn2019weakly}, we first generate CAMs for training images by adopting the method of \cite{zhou2016learning} using a multi-label image classification network. For a fair comparison, we use a ResNet-50~\cite{he2016deep} model as the classification network, as used in other CAM-based methods~\cite{ahn2018learning,ahn2019weakly,huang2018weakly,yao2020saliency}. We directly utilize the raw CAMs to generate pseudo-labels by thresholding their confidence scores for each class label at every pixel predicted to be an object by the class agnostic objectness network (see Fig.~\ref{fig:pseudo}(B)). We can formalize this procedure as follows:
\begin{gather} \label{eq:cam}
 \resizebox{.5\hsize}{!}{$\mathcal{G}^{_{\mathcal{C}_m}}_{i,j} = \begin{cases} 
      \underset{k\in K}{\arg\max}(\mathcal{C}_{m}(i,j,k)) & \text{if} \hspace{0.08cm} \mathcal{O}_{i,j} > 0 \hspace{0.1cm} \hspace{0.1cm} \mathcal{C}_m(i,j) > \delta \\
      0 &  \text{otherwise}
   \end{cases}\hspace{0.1cm},$}
\end{gather}
\noindent where $\mathcal{G}^{_{\mathcal{C}_m}}_{i,j}$ denotes the pseudo-label value at pixel $(i,j)$, $K$ is the set of class indices, $\mathcal{O}_{i,j}$ is the objectness score, $\mathcal{C}_m$ is the non-thresholded CAM proposals, and $\delta$ is a threshold (we use $\delta=0.01$ in all experiments).\\

\noindent \textbf{Box-Driven Pseudo-labels.}\label{sec:label_box}
The simplest box-driven pseudo-labels can be obtained by filling the bounding box annotations with the corresponding class label. Some methods~\cite{khoreva2017simple,song2019box} use semi-automatic segmentation techniques (e.g., CRF~\cite{papandreou2015weakly}, GrabCut~\cite{rother2004grabcut}) to further refine the box annotations, as rectangular regions contain a significant number of incorrectly labeled background pixels. However, these techniques are time consuming and the quality of the pseudo-label is lacking. To address this challenge, we propose an approach to generate pseudo-labels using the class agnostic objectness masks, $\mathcal{O}$, and the box annotations, $\mathcal{B}$.
		
Following common practice~\cite{khoreva2017simple,ibrahim2018weakly,song2019box}, if any two bounding boxes overlap, we assume the box with smaller area appears in front. Additionally, if the overlap between any box and the largest box in the image is greater than some threshold, we only keep the inner 60\% of the box and fill the rest of the box as 255 (which is ignored during training). The intuition behind the \textit{ignoring strategy} is simply trading-off lower recall (ignore more pixels where high-degree of overlap occurs) for higher precision (more pixels are correctly labelled). 
We then mask the resulting box proposal, $\mathcal{B}$, with the objectness map, $\mathcal{O}$, to filter out the irrelevant regions from $\mathcal{B}$ and $\mathcal{O}$, and only keep the regions overlapping the object of interest. We set any pixel to the background class if it does not overlap any boxes. Formally, for each bounding box, $\mathcal{B}_k$, $k \in \{1,...,n\}$, in an image:
\begin{gather}\label{eq:box}
   \resizebox{.55\hsize}{!}{$\mathcal{G}^{_{\mathcal{B}}}_{_{\text{ign}}}(i,j) = \begin{cases} 
   \mathcal{B}^{\text{cls}}_k & \text{if} \hspace{0.1cm} \mathcal{O}_{i,j} > 0, \hspace{0.1cm}\mathcal{B}_{0}\cap\mathcal{B}_k < \alpha ,\hspace{0.1cm} (i,j) \in \mathcal{B}_k \\
       \mathcal{B}^{\text{cls}}_k & \text{if} \hspace{0.1cm} \mathcal{O}_{i,j} > 0, \hspace{0.1cm} \mathcal{B}_0\cap\mathcal{B}_k > \alpha , \hspace{0.1cm} (i,j) \in \mathcal{B}^{\text{in}}_k \\
       255 & \text{if} \hspace{0.1cm} \mathcal{O}_{i,j} > 0, \hspace{0.1cm} \mathcal{B}_0\cap\mathcal{B}_k > \alpha   \hspace{0.1cm} (i,j) \in \mathcal{B}^{\text{out}}_k\\
      0 &  \text{otherwise}
   \end{cases} \hspace{0.2cm},$}
\end{gather}
		

\noindent where $\mathcal{B}_0$ denotes the largest box, $n$ is the number of boxes in each image, $\mathcal{B}^{\text{out}}_k$ is the outer 40\% of the bounding box's area, $\mathcal{B}^{\text{in}}_k$ is the inner 60\% of the bounding box, `$\cap$' calculates the area of intersection between two bounding boxes, and $\alpha$ is a threshold (we set $\alpha=0.3$). 
\vspace{-0.1cm}
\subsection{Semantic Objectness Network: SONet}\label{sec:net}
\begin{wrapfigure} {r}{6.2cm}
\vspace{-0.75cm}
\begin{center}
		\includegraphics[width=0.47\textwidth]{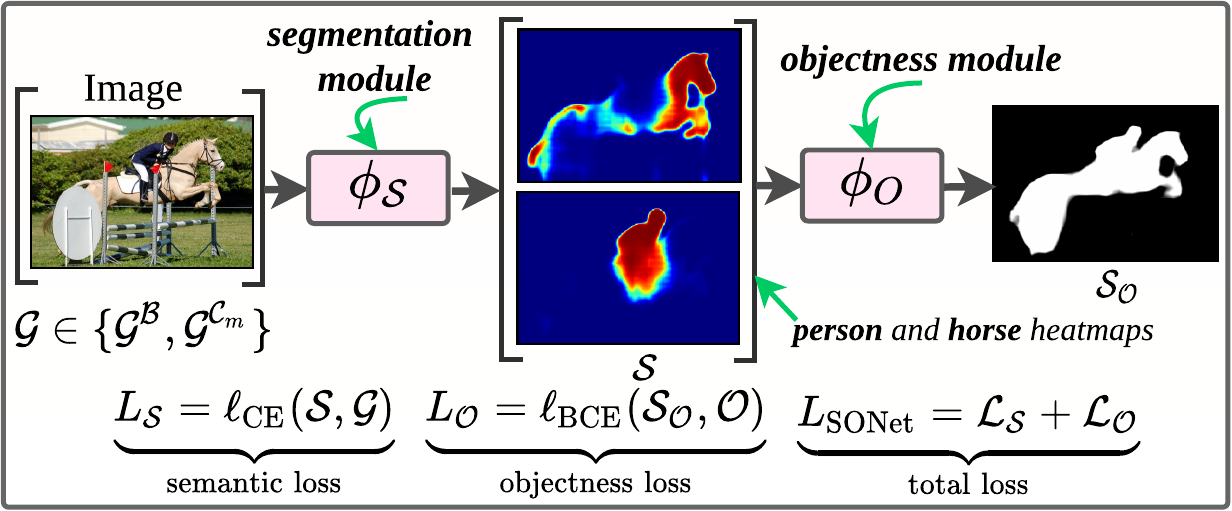}
	    \vspace{-0.15cm}
		\caption{Overview of our proposed SONet. With the generated pseudo-labels, we jointly train SONet on the target dataset, $\mathcal{D_T}$, to predict semantics, $\mathcal{S}$, and objectness, $\mathcal{S_O}$.}
		\label{fig:network}
	\end{center}
	\vspace{-0.6cm}
\end{wrapfigure}

The Semantic Objectness Network (SONet) consists of a segmentation network and an objectness module. The objectness module receives the output of the segmentation network as input, and predicts a binary mask (see Fig.~\ref{fig:network}). \\
\noindent \textbf{Network Architecture.} \blue{We use DeepLabv3~\cite{chen2017rethinking} as our segmentation network, $\phi_\mathcal{S}$, which outputs feature maps of 1/16 of the input image size. Given an input image, $\mathcal{I}_t$, $\phi_\mathcal{S}$ generates a semantic segmentation map, $\mathcal{S}$, using the pseudo-label as supervision. The objectness module, $\mathcal{\phi_O}$, takes $\mathcal{S}$ as input and consists of a stack of five convolutional layers that includes batch normalization and ReLU layers (see Table S1 in the supplementary for architectural details). We use a 3$\times$3 kernel in the first four convolution layers and use a 1$\times$1 kernel in the last layer which outputs the objectness map, $\mathcal{S_O}$. The procedure for obtaining the semantic and objectness maps can be described as:
\begin{gather}
  \mathcal{S} = \phi_\mathcal{S}(\mathcal{I}_t; \textbf{W}_{T}), \hspace{0.3cm} \mathcal{S_O} = \phi_O (\mathcal{S};\textbf{ W}_{O}),
\end{gather}
where $\textbf{W}_{T}$ and $\textbf{W}_{O}$ refer to trainable weights for the $\phi_\mathcal{S}$ and $\phi_O$ modules, respectively. } \\

\noindent\textbf{Joint Learning of Semantics and Objectness.} \blue{We train our proposed SONet method using the generated pixel-level semantic and objectness pseudo-labels in an end-to-end manner (see Fig.~\ref{fig:network}). Let $\mathcal{D_T} = (\mathcal{I}_t, \mathcal{G}, \mathcal{O})$ denote the target semantic segmentation dataset with images $\mathcal{I}_t$, pseudo-labels $ \mathcal{G}\in \{\mathcal{G}^{_{\mathcal{C}_m}},\mathcal{G}^{_{\mathcal{B}}}\}$, and $\mathcal{O}\in\{0, 1\}$ is the objectness prior. More specifically, let $\mathcal{I}_t\in\mathbb{R}^{h\times w\times 3}$ be a training image from $\mathcal{D_T}$ with semantic segmentation pseudo-label $ \mathcal{G}\in\mathbb{R}^{h\times w}$ and the objectness prior $ \mathcal{O}\in\mathbb{R}^{h\times w}$. We denote the pixel-wise cross entropy loss function $L_\mathcal{S}$ and $L_\mathcal{O}$ between ($\mathcal{S}, \mathcal{G}$) and $(\mathcal{S_O}, \mathcal{O})$, respectively. The final loss function of the network is the sum of the segmentation and objectness losses as follows: 
\begin{gather}
 L_\mathcal{S} = \ell_{\text{CE}} (\mathcal{S},  \mathcal{G}), \hspace{0.15cm} L_\mathcal{O}  = \ell_{\text{BCE}}  (\mathcal{S_O}, \mathcal{O}), \hspace{0.15cm}
  L_{\text{SONet}} = L_\mathcal{S} + L_\mathcal{O},
\end{gather}
where $\ell_{\text{CE}}$ and $\ell_{\text{BCE}}$ denote the multi-class and binary cross entropy loss function, respectively. The joint training allows our network to propagate objectness information together with semantics, and suppress erroneous decisions which allows for more accurate final predictions for both outputs. During inference, we simply take the segmentation map to measure the overall performance of our proposed approach.}
 
\vspace{-0.2cm}
\section{Experiments}\label{sec:exp}
We evaluate our proposed framework on the PASCAL VOC 2012~\cite{everingham2015pascal} and Cityscapes~\cite{cordts2016cityscapes} semantic segmentation benchmarks. \blue{We generate objectness masks for the VOC12 target dataset from two different objectness trained models on COCO-Stuff: \textit{overlapping} (all images) and \textit{non-overlapping} (images with no overlapping objects with \textit{target} data)}.
We also report experiments under \textit{domain adaptation} settings by training on a different target dataset, OpenV5~\cite{kuznetsova2018open}, and evaluating on VOC12. OpenV5~\cite{kuznetsova2018open} is a recently released dataset consisting of image-level, bounding box, and semantic segmentation annotations for over 600 classes. For these experiments, we randomly select 42,621 images from the same 21 classes as VOC12 and generate pseudo-labels using our box-driven approach. We train SONet with the generated pseudo-labels from OpenV5 and then evaluate on VOC12 (denoted as O $\rightarrow$ V). We also finetune SONet on VOC12 before evaluation (denoted as O + V). 
We report experimental results with different backbone networks for a fair comparison. 
\vspace{-0.2cm}
\subsection{Analysis of Generated Pseudo-labels}\label{sec:label_eval}
We first evaluate the quality of our generated pseudo-labels to explore the upper bound for different types of weak supervision and report the results in Table~\ref{tab:voc12_val_1}(a). We consider the generated pseudo-labels as predictions and obtain the upper bound for each supervision type by calculating the mIoU between the pseudo-label and the groundtruth. To generate CAMs pseudo-labels, $\mathcal{C}_m$, we simply threshold the scores of raw CAMs. When we apply our objectness mask, $\mathcal{O}$, to improve the boundary of CAMs ($\mathcal{G}^{\mathcal{C}_m}$), we obtain 70.6\% mIoU. Further, using the bounding boxes and objectness map ($\mathcal{G}^{_{\mathcal{B}}}$) achieves 76.6\% mIoU that further improves the upper bound mIoU by 6\%. In addition, applying the non-overlapping objectness mask, $\mathcal{O_N}$, substantially improves the CAMs ($\mathcal{G}_{\dagger}^{\mathcal{C}_m}$) or bounding box ($\mathcal{G}_{\dagger}^{_{\mathcal{B}}}$) proposals. As shown in Table~\ref{tab:voc12_val_1}(a), exploiting   
\begin{wraptable}{r}{7.4cm}

	\vspace{-0.4cm}
	
	\begin{center}
		\def\arraystretch{1.1}
		\resizebox{0.58\textwidth}{!}{
			\begin{tabular}{ c|  c}
				
                \specialrule{1.2pt}{1pt}{1pt}
                Sup. &  Train \\
                \specialrule{1.2pt}{1pt}{1pt}
                 $\mathcal{C}_m$ & 48.3 \\ 
				  
				   $\mathcal{G}_{\dagger}^{\mathcal{C}_m}$ & 63.5 \\ 
				   $\mathcal{G}^{\mathcal{C}_m}$ & 70.6 \\ 
				   \hline
				 $\mathcal{B}$ & 60.2 \\ 
				  
				  $\mathcal{G}_{\dagger}^{_{\mathcal{B}}}$  & 68.6 \\
				  
				  $\mathcal{G}^{_{\mathcal{B}}}$  & 76.6 \\ 
				\specialrule{1.2pt}{1pt}{1pt}
				\multicolumn{2}{c}{(a)} \\
		
			\end{tabular}
			\hspace{0.2cm}
			
			\begin{tabular}{c|c |c}
				\specialrule{1.2pt}{1pt}{1pt}
				Method &  \multicolumn{1}{c|}{Sup.} & Val. \\
				\specialrule{1.2pt}{1pt}{1pt}
			    \multirow{6}{*}{SONet} & $\mathcal{C}_m$ &  50.2\\
			    & $\mathcal{G}_{\dagger}^{\mathcal{C}_m}$ &  65.3\\
				& $\mathcal{G}^{\mathcal{C}_m}$ &  \textbf{70.5}\\
				\cline{2-3}
				& $\mathcal{B}$ &  54.6\\
				& $\mathcal{G}_{\dagger}^{_{\mathcal{B}}}$  & 67.9 \\
				& $\mathcal{G}^{_{\mathcal{B}}}$ &  \textbf{73.8}\\
				\specialrule{1.2pt}{1pt}{1pt}
                \multicolumn{3}{c}{(b)} \\

			\end{tabular}
			\hspace{0.2cm}
			\begin{tabular}{c|c |c}
				\specialrule{1.2pt}{1pt}{1pt}
				Method &  Sup. & mIoU \\
				\specialrule{1.2pt}{1pt}{1pt}
				\rowcolor{maroon!7}
				\multicolumn{3}{c}{\textbf{O $\rightarrow$ V}}  \\
				\multirow{2}{*}{SONet} & $\mathcal{B}$ & 51.5 \\
				 & $\mathcal{G}^{_{\mathcal{B}}}$ &  71.0\\
				 \rowcolor{maroon!7}
				\multicolumn{3}{c}{\textbf{O + V}}  \\
				 \multirow{2}{*}{SONet} & $\mathcal{G}^{_{\mathcal{B}}}$ & 75.9 \\ 
				 & $\mathcal{G}^{_{\mathcal{B}}}_{_{\text{ign}}}$ & 76.9 \\
				\specialrule{1.2pt}{1pt}{1pt}
				\multicolumn{3}{c}{(c)} \\
				
			\end{tabular}

			}
			\vspace{-0.4cm}
			\caption{ (a) Upper-bound analysis for different pseudo-label types on the VOC12 train set. (b) Quantitative results on VOC12 val set for the baseline and our approach. (c) SONet's performance under domain adaptation settings, trained on OpenV5 and then either directly evaluated (O $\rightarrow$ V) or fine-tuned (O + V) on VOC12.}
	        
			\label{tab:voc12_val_1}
		\end{center}
		\vspace{-0.7cm}
\end{wraptable}
an objectness map with CAMs or bounding boxes significantly improves the quality of pseudo-labels as it removes incorrectly labeled pixels from the CAM and bounding box proposals. 
Next, we evaluate the performance of our proposed SONet (Table~\ref{tab:voc12_val_1}(b)) with CAMs, box annotations, and the generated pseudo-labels. SONet trained with box proposals achieves 54.6\% mIoU which outperforms the same model trained using CAM proposals (50.2\%). When we use our generated pseudo-labels during training, SONet achieves 70.5\% mIoU ($\mathcal{G}^{\mathcal{C}_m}$) and 73.8\% ($\mathcal{G}^{_{\mathcal{B}}}$) on the VOC12 val set. 
In the domain adaptation settings (see Table~\ref{tab:voc12_val_1}(c)), SONet trained only with OpenV5 groundtruth boxes achieves 51.5\% mIoU accuracy on the VOC12 val set. When SONet is trained on OpenV5 with $\mathcal{G}^{_{\mathcal{B}}}$ as supervision, it drastically improves the overall mIoU to 71.0\% (note that in this setting we \textit{only} use OpenV5 images to train SONet). Additionally, fine-tuning SONet on the VOC12 training set with $\mathcal{G}^{_{\mathcal{B}}}$ supervision increases the mIoU to 76.9\%. 
These experiments indicate that our pseudo-label generation technique achieves good upper bound performance compared to the groundtruth. 
\subsection{Image Segmentation Results}\label{sec:img_seg}
\begin{wraptable}{r}{7.0cm}
     \vspace{-0.8cm}

	    \begin{center}

		
		\resizebox{0.52\textwidth}{!}{

		\begin{tabular}{l|c|c | c| c}
				\specialrule{1.2pt}{1pt}{1pt}
				\multirow{2}{*}{Method} & \multirow{2}{*}{Backbone}  & \multirow{2}{*}{Guidance} & \multicolumn{2}{c}{\textbf{mIoU}} \\
				\cline{4-5} 
				&&&val & test \\
				
			\specialrule{1.2pt}{1pt}{1pt}
				
				

				\rowcolor{maroon!7}
				\multicolumn{5}{c}{\textit{Weakly-Supervised Approaches}}  \\
				
				\rowcolor{maroon!7}
				\multicolumn{5}{l}{Image-Level Supervision (CAM)}  \\
				FlickleNet~\cite{lee2019ficklenet}  & Res-101 &Saliency & 64.9 &65.3 \\
				
				OAA$^*$~\cite{jiang2019integral} &  Res-101 & Saliency &65.2& 66.4\\
				ME~\cite{fan2020employing}  & Res101 &Saliency & 67.2 &66.7 \\
				
				ICD~\cite{fan2020learning}  & Res101 &Saliency & 67.8 &68.0 \\
				SGAN$^*$ & Res-101 & Saliency &67.1& 67.2\\
				\midrule
				\textbf{SONet-O}$^*$ &Res-101 & Objectness &64.5& 65.8\\
				
				\textbf{SONet$^*$} &Res-101 & Objectness &\textbf{68.1}& \textbf{69.7}\\
				\textbf{SONet} & Res-101  & Objectness &\textbf{70.5}& \textbf{71.5}\\

				\hline
				\rowcolor{maroon!7}
				\multicolumn{5}{l}{Box-Level Supervision}  \\
				
				
                
                
                
                
				
				SDI$^*$~\cite{khoreva2017simple} & Res-101 & BSDS &69.4 &- \\ 
				BCM$^*$~\cite{song2019box}& Res-101 &  CRF &70.2 & - \\

				\hline
				
			
			      \textbf{SONet}$^*$ & Res-101 & Objectness &\textbf{72.2}& 73.7\\
			      \textbf{SONet} & Res-101 & Objectness &\textbf{74.8}& 76.0\\
			
				\hline
					\rowcolor{maroon!7}
				\multicolumn{5}{c}{\textit{Semi-Supervised Approaches}}  \\
				
				WSSL~\cite{papandreou2015weakly} & VGG-16 &1.4k GT &64.6 &-\\
                SDI~\cite{khoreva2017simple}  & VGG-16 &1.4k GT  &65.8& 66.9\\
                FickleNet~\cite{lee2019ficklenet} &VGG-16 &1.4k GT   &65.8&-\\
                \textbf{SONet} & VGG-16 & - & \textbf{66.1} & \textbf{67}\\
                
				\specialrule{1.2pt}{1pt}{1pt}
             
			\end{tabular}}
			\caption{Quantitative comparison with weakly and semi supervised methods on the PASCAL VOC12 validation and test sets. SONet-O denotes the method used non-overlapping objectness prior. Methods marked by $*$ used DeepLabv2-Res101.}
			\label{tab:voc12_val_2}
	    \end{center}
	    \vspace{-0.85cm}
\end{wraptable} 
In this section, we compare our proposed SONet method with previous state-of-the-art WSSS and SSSS methods~\cite{papandreou2015weakly,khoreva2017simple,huang2018weakly,song2019box,lee2019ficklenet,zeng2019joint,jiang2019integral,fan2020employing,fan2020learning,chang2020weakly}. Table~\ref{tab:voc12_val_2} presents a comparison with recent weakly and semi supervised methods using image and bounding box-level supervision. For fair comparison in WSSS setting, we compare with other methods that use ResNet-101 as the backbone and additional guidance (e.g., saliency maps and optical flow) as supervision. SONet$^*$ outperforms the current state-of-the-art image-level + extra guidance based methods by a reasonable margin, achieving 68.1\% mIoU on the VOC12 val set. Interestingly, SONet-O$^*$, which is trained on the pseudo-labels generated under the non-overlapping settings, also achieves comparable performance with the baseline WSSS methods. When compared to methods that use bounding box-level supervision with extra guidance, SONet$^*$ also improves upon the state-of-the-art~\cite{khoreva2017simple, song2019box} by 2.0\%. Note that both BCM~\cite{song2019box} and SDI~\cite{khoreva2017simple} take much longer to produce pseudo-labels than our approach due to their iterative procedures and use complex training protocols. We do not include results for a recent box based method, Box2Seg~\cite{kulharia2020box2seg}, as they use a higher capacity network architecture~\cite{xiao2018unified} for segmentation learning without publicly available code. Our SONet method achieves 74.8\% mIoU on the VOC12 val set which is very close (2.4\% lower) to the fully supervised trained baseline~\cite{chen2017rethinking} model (77.2\%). In the SSSS setting, we use a similar backbone network as existing methods to ensure a fair comparison. Note, existing SSSS methods use a portion of the target semantic segmentation groundtruth while we \textit{solely} use our generated pseudo-labels to train the network. Surprisingly, SONet (VGG-16 backbone) marginally outperforms the existing SSSS methods (66.1\% vs. 65.8\% mIoU). These results demonstrate that our pseudo-label generation procedure is flexible and achieves substantial improvements or competitive performance compared to existing methods in WSSS and SSSS. 
\begin{table*} [t]
   
    \begin{center}
    \resizebox{1.\textwidth}{!}{%
    \def\arraystretch{1.1}
    \setlength{\tabcolsep}{3.2pt}
    \begin{tabular}{l|c|cccccccccccccccccccc|c}

         \multicolumn{1}{c|}{Method}&\rotatebox{90}{Sup.}&\rotatebox{90}{aero}&\rotatebox{90}{bike}&\rotatebox{90}{bird}&\rotatebox{90}{boat}&\rotatebox{90}{bottle}&\rotatebox{90}{bus}&\rotatebox{90}{car}&\rotatebox{90}{cat}&\rotatebox{90}{chair}&\rotatebox{90}{cow}&\rotatebox{90}{table} &\rotatebox{90}{dog}&\rotatebox{90}{horse}&\rotatebox{90}{mbike}&\rotatebox{90}{person}&\rotatebox{90}{plant}&\rotatebox{90}{sheep}&\rotatebox{90}{sofa}&\rotatebox{90}{train}&\rotatebox{90}{tv}&\rotatebox{90}{\textbf{mIoU}}\\
      \specialrule{1.2pt}{1pt}{1pt}
       
       
       
       
       
        
         
         SONet & $\mathcal{G}^{\mathcal{C}_m}$ &84.4&37.6&83.7&63.9&51.6&88.4&84.7&68.3&30.8&81.4&57.9&68.0&79.6&83.6&74.2&58.5&84.6&53.4&81.5&57.1&69.7 \\
        \midrule 
       
        SONet(O$\rightarrow$V) & $\mathcal{G}^{\mathcal{B}}$ & 91.3 &39.9&89.7&68.5&68.6&89.8&77.0&80.8&21.4&71.9&34.2&81.3&83.4&82.4&71.4&58.6&82.9&53.2&\textbf{86.4}&66.5& 71.1\\
        
         SONet & $\mathcal{G}^{\mathcal{B}}$&   91.4&37.3&88.7&68.0&66.0&94.1&88.0&79.9&32.3&83.4&64.3&77.5&86.3&78.0&74.4&59.4&86.3&57.3&84.8&66.2& 74.1\\
         
        SONet(O+V)& $\mathcal{G}^{\mathcal{B}}$ &90.7&40.0 &90.2&69.7&72.7&94.1&87.4&79.2&32.7&\textbf{86.7}&62.6&80.1&\textbf{88.5}&81.3&74.2&62.6&\textbf{91.9}&\textbf{58.5}&\textbf{89.2}&69.5& 76.0 \\
         
          SONet & $\mathcal{G}^{\mathcal{B}}_{\text{ign}}$ & 92.1 & 40.9 & 90.6 & 68.5 & \textbf{74.0} &94.1& 87.1 & 83.2 & 31.3 & \textbf{86.4} & \textbf{67.2} & 78.2 & 84.6 & 84.0 & 77.1 & 61.6 & \textbf{90.6} & 55.3 & \textbf{85.4} & 69.2& 76.0 \\
          
         SONet (O+V)& $\mathcal{G}^{\mathcal{B}}_{\text{ign}}$&  91.8&39.9&89.9&\textbf{71.3}&\textbf{74.8}&\textbf{94.6}&\textbf{88.2}&80.9&33.0&\textbf{89.5}&62.8&82.5&\textbf{89.7}&83.8&76.9&62.8&\textbf{90.3}&\textbf{59.9}&\textbf{89.3}&70.0& 77.0 \\

       \midrule 
        DeepLabV3& $\mathcal{F}$& \textbf{92.9} & \textbf{60.3} & \textbf{93.0} & 70.5 & 73.3 & 94.1 & 88.1 &\textbf{90.9} & \textbf{35.3} & 83.4 & 65.7 & \textbf{86.3} & 87.5 & \textbf{85.2} & \textbf{86.5} & \textbf{63.8} & 88.1 & 57.6 & 85.0 & \textbf{72.3} &\textbf{78.8} \\

      \specialrule{1.2pt}{1pt}{1pt}
    \end{tabular}}
    \caption{Class-wise IoU on the VOC12 test set. (O $\rightarrow$ V) refers to training on OpenV5 and test on VOC. (O + V) denotes training on OpenV5 and fine-tuning on VOC12. $\mathcal{F}$ denotes full supervision. Bolded values denote the results that surpass the fully supervised method.}
  \label{tab:quant_pascal_test}
   \end{center}
  \vspace{-0.7cm}
\end{table*}
Table~\ref{tab:quant_pascal_test} presents a class-wise IoU comparison of SONet with different training strategies as well as with the fully supervised baseline DeepLabv3 model on the VOC12 test set. Notably, although the fully supervised model achieves the highest mIoU, SONet (O+V) trained using $\mathcal{G}^{_{\mathcal{B}}}_{_{\text{ign}}}$ outperforms the fully supervised model on \textit{half} of the categories, and is competitive in many others. In general, when trained using bounding box-based pseudo-labels, SONet performs well on rectangular shaped classes, e.g., bus, car, tv, cow, bottle, bus, and train. However, it is still difficult for any training protocol combined with SONet to achieve comparable performance with classes like bike, motorbike, cat, dog or person, where the objects have complex boundary information or are occluded with other classes, e.g., person on a horse or bike. Furthermore, using the ignore strategy ($\mathcal{G}^{_{\mathcal{B}}}_{_{\text{ign}}}$ vs. $\mathcal{G}^{_{\mathcal{B}}}$) improves the performance notably for both the normal and domain adaptation settings. The quantitative results indicate that our SONet model can achieve competitive performance with the fully supervised model, showing the effectiveness of the proposed pseudo-label generation and the joint learning techniques.




             
\begin{wraptable}{r}{4.1cm}
     \vspace{-0.7cm}

	    \begin{center}

		\setlength \tabcolsep{4.1pt}
		\resizebox{0.31\textwidth}{!}{

		\begin{tabular}{l c c c}
				\specialrule{1.2pt}{1pt}{1pt}
				    Method &  Sup. & mIoU \\
				\specialrule{1.2pt}{1pt}{1pt}
				DLabv3 (Things) & Full & 81.5  \\
				SONet (Things) & $\mathcal{G}^B$ & 76.6\\
				\specialrule{1.2pt}{1pt}{1pt}
             
			\end{tabular}}
			\caption{Quantitative results on the Cityscapes~\cite{cordts2016cityscapes} val set.}
			\label{tab:cityscapes}
	    \end{center}
	    \vspace{-0.65cm}
\end{wraptable} 
\blue{We further use Cityscapes~\cite{cordts2016cityscapes} as our target dataset and report results in Table~\ref{tab:cityscapes}. Cityscapes consists of eight \textit{things} classes and 11 stuff classes. Similar to the VOC12, we first generate class agnostic objectness masks for the Cityscapes train set and combine it with the bounding box annotations to generate semantic pseudo-labels. Since our objectness network is not trained for generating masks for \textit{stuff} classes, we only consider the \textit{things} classes from Cityscapes during pseudo-label generation, training, and evaluation. Next, we train DeepLabv3-ResNet50~\cite{chen2017rethinking} with full supervision as a baseline and SONet (DeepLabv3-ResNet50 as backbone) using the generated pseudo-labels ($\mathcal{G}^B$).
Table~\ref{tab:cityscapes} shows that, 
our SONet performs well (76.6\% mIoU) and obtains 94\% relative to the baseline (similar to our results on VOC12). This result further confirms the generalizability of our pseudo-label generation technique despite the significant distribution gap between the target (Cityscapes~\cite{cordts2016cityscapes}) and the source (COCOStuff~\cite{caesar2018coco}) dataset.}

\vspace{-0.2cm}
\subsection{Ablation Studies}\label{sec:SONet_abl} 

\noindent\textbf{Effectiveness of Objectness Branch. 
}
We first validate the effect of the objectness branch by comparing the results of SONet trained in both multi- and single-task settings. We train SONet with and without the objectness branch. Note that SONet without the objectness branch is equivalent to DeepLabv3~\cite{chen2018deeplab}. The result of these comparisons are summarized in Table~\ref{tab:voc12_abl} (a). It is clear that the models trained with the objectness branch achieve superior performance compared to the models trained only for the task of semantic segmentation. Interestingly, the objectness branch improves the boundary details to bring more smoothness (see Fig. in Table~\ref{tab:voc12_abl} (top row)) as expected, as well as the semantic information (see the figure in Table~\ref{tab:voc12_abl} (bottom row)). SONet's multi-task objective not only provides it with the ability to robustly predict both binary and semantic segmentation, but the objectness-based learning naturally provides the segmentation network a significant performance boost.\\

\noindent \textbf{Effectiveness of Ignoring Strategy.} In Table~\ref{tab:voc12_abl}(b), we compare our ignore strategy to the strategy in SDI~\cite{khoreva2017simple} when trained using SONet. We show that our ignoring strategy outperforms both SDI~\cite{khoreva2017simple} and SONet trained without an ignore strategy.\\

\begin{table}[t]
        \centering
		
		\def\arraystretch{1.2}
			\setlength\tabcolsep{3.6pt}
		\resizebox{0.95\textwidth}{!}{
		    \begin{tabular}{c| c c | c c}
				
 				\specialrule{1.2pt}{1pt}{1pt}
                
                 \multirow{2}{*}{ \textbf{Sup.}} & \multicolumn{2}{c|}{\textbf{VOC12}} & \multicolumn{2}{c}{\textbf{O $\rightarrow$ V}}\\
                 \cline{2-5}
 				& \cite{chen2018deeplab}& SONet & \cite{chen2018deeplab} & SONet  \\
 				\specialrule{1.2pt}{1pt}{1pt}
 				
 				 $\mathcal{B}$ & 52.1 & \textbf{54.6} & 50.5 & \textbf{51.5} \\
 				 $\mathcal{G^B}$ & 73.1 & \textbf{73.8} & 70.4& \textbf{71.0} \\
				
 				\specialrule{1.2pt}{1pt}{1pt}
 				\multicolumn{5}{c}{(a)} \\
 			\end{tabular}
 			\hspace{0.2cm}
 			\begin{tabular}{c|c|c}
				 Sup. 
				& Val. & Test \\
				\specialrule{1.2pt}{1pt}{1pt}
			 ~\cite{khoreva2017simple}${_{\text{ign}=0.1}}$ & 67.9 & - \\ 
				 ~\cite{khoreva2017simple}${_{\text{ign=0.2}}}$ & 67.6 & -\\
				\midrule
				 $\mathcal{G}^{_{\mathcal{B}}}$ &  73.8 & 74.1 \\
				 $\mathcal{G}^{_{\mathcal{B}}}_{_{\text{ign}}}$ & 74.8 & 76.0 \\
				\specialrule{1.2pt}{1pt}{1pt}
				\multicolumn{3}{c}{(b)} \\

			\end{tabular}
			\hspace{0.2cm}
			\setlength\tabcolsep{0.2pt}
					    \begin{tabular}{*{5}{c}}
			\includegraphics[width=0.12\textwidth]{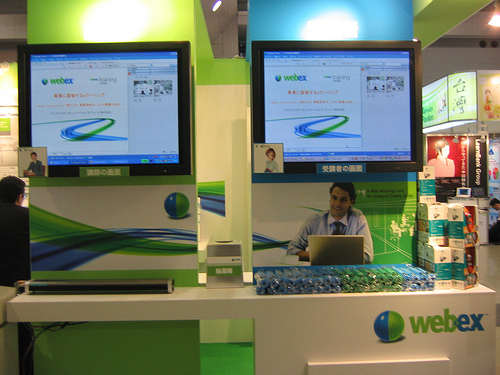}&
	       \includegraphics[width=0.12\textwidth]{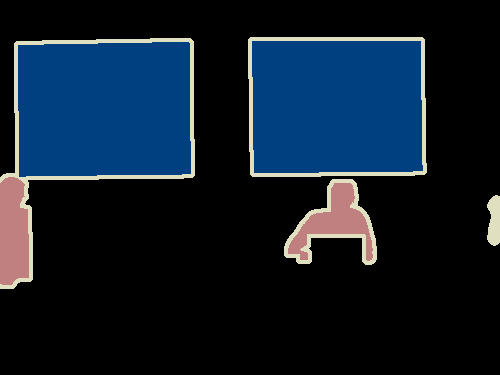}&		\includegraphics[width=0.12\textwidth]{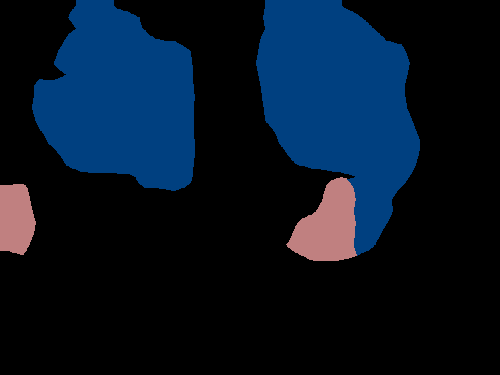}&
	       \includegraphics[width=0.12\textwidth]{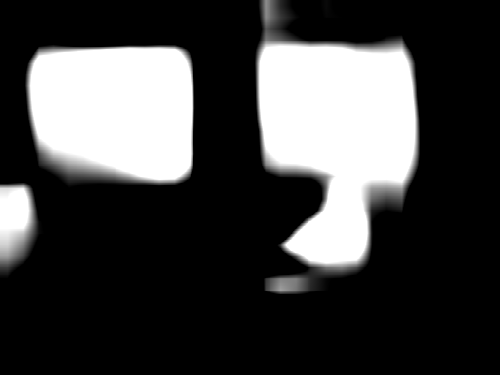}&
	      \includegraphics[width=0.12\textwidth]{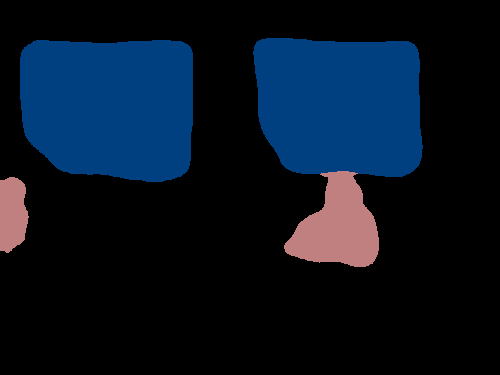}\\

	   		\includegraphics[width=0.12\textwidth]{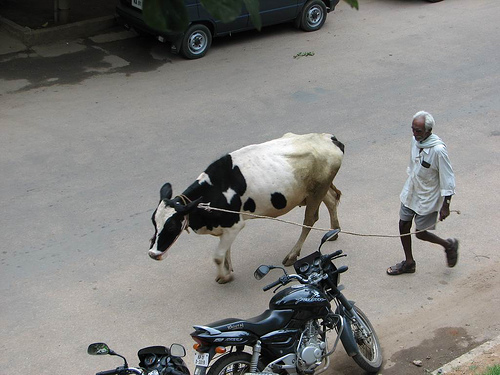}&
	       \includegraphics[width=0.12\textwidth]{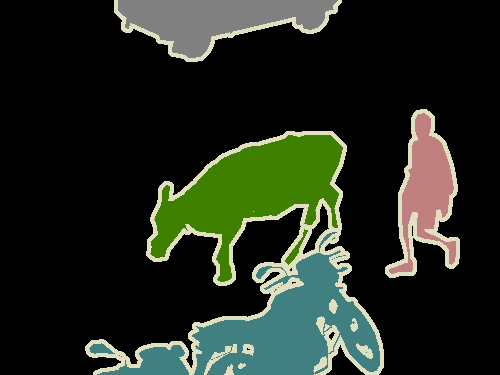}&		
	       \includegraphics[width=0.12\textwidth]{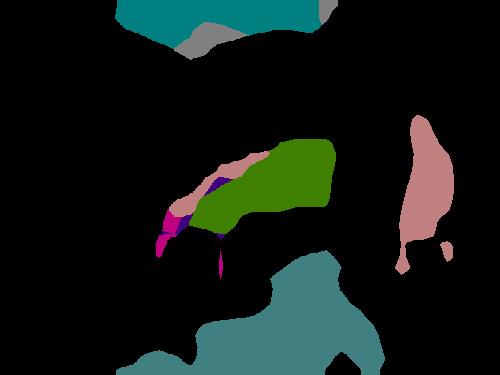}&
	       \includegraphics[width=0.12\textwidth]{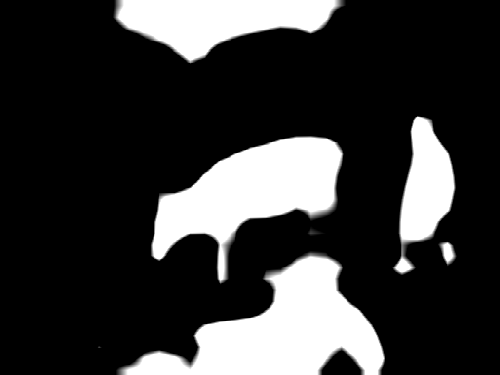}&
	      \includegraphics[width=0.12\textwidth]{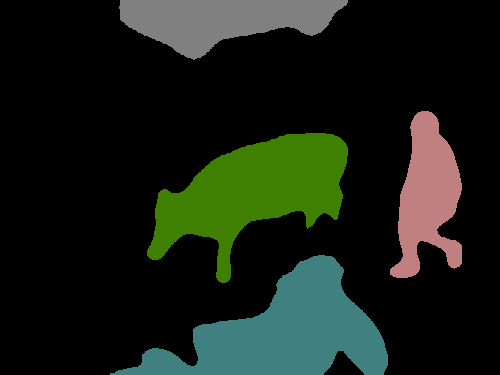}\\
		  Image & GT & DeepLabv3 & Mask & SONet\\	
			\end{tabular}
		}
		\caption{ (a) Comparison between SONet and DeepLabv3~\cite{chen2018deeplab} on the VOC12 val set. (b) Comparison of our ignore strategy with SDI~\cite{khoreva2017simple}. (Right) Visualization of the effect on segmentation results of the model trained with or w/o objectness branch.} 
	\label{tab:voc12_abl}
	\vspace{-0.5cm}
\end{table}

\begin{figure}[t]
    	\begin{center}
		\setlength\tabcolsep{0.4pt}
		\def\arraystretch{0.2}
		\resizebox{0.95\textwidth}{!}{

		  \begin{tabular}{*{4}{c}}		
			\centering	
            \includegraphics[width=0.13\textwidth]{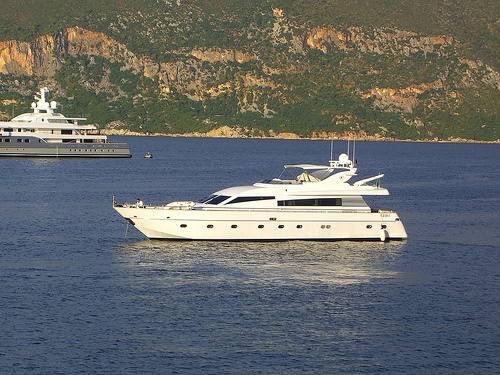}&
            \includegraphics[width=0.13\textwidth]{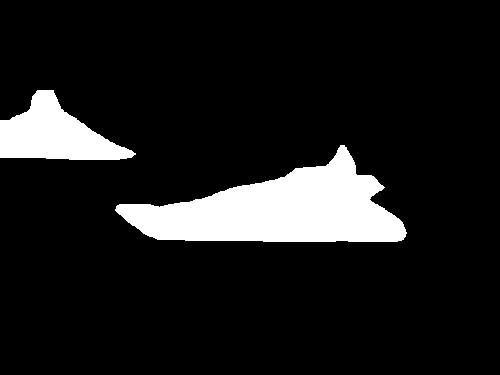}&
            \includegraphics[width=0.13\textwidth]{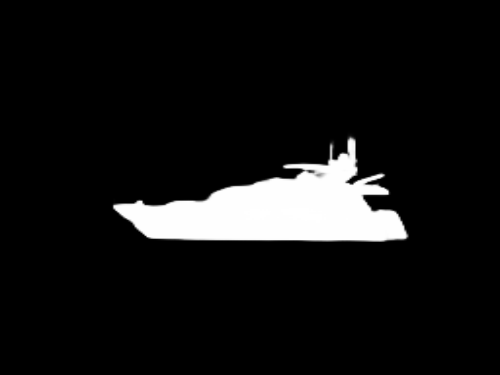}&
            \includegraphics[width=0.13\textwidth]{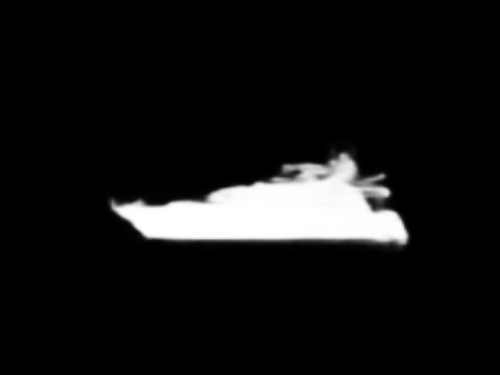}\\
            
           \includegraphics[width=0.13\textwidth]{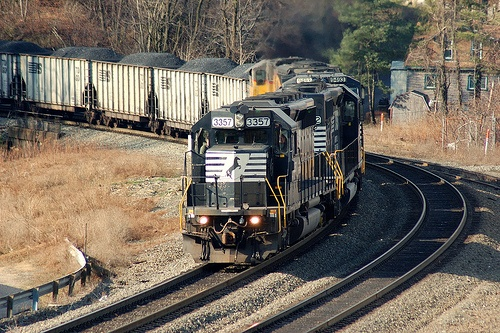}&
           \includegraphics[width=0.13\textwidth]{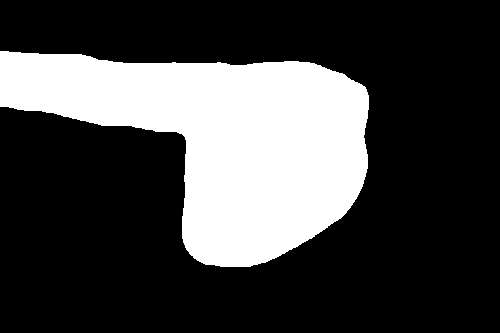}&
             \includegraphics[width=0.13\textwidth]{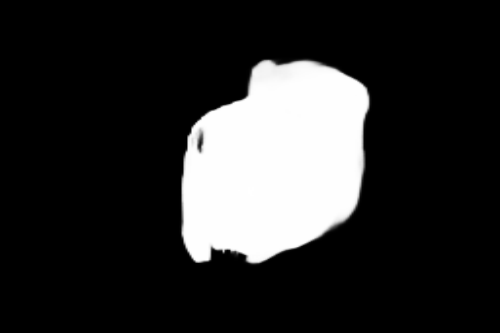}&
            \includegraphics[width=0.13\textwidth]{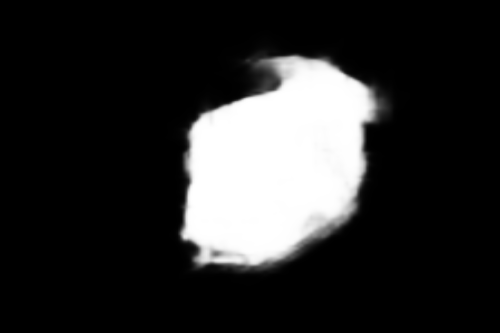}\\
            
            \small{Image} & Ours & \small{BASNet~\cite{qin2019basnet}} & \small{PicaNet~\cite{liu2018picanet}}\\

	        \multicolumn{4}{c}{(a)} \\

			\end{tabular}
			\hspace{0.04cm}
			
			\begin{tabular}{*{2}{c}}		
			\centering	
            \includegraphics[width=0.17\textwidth]{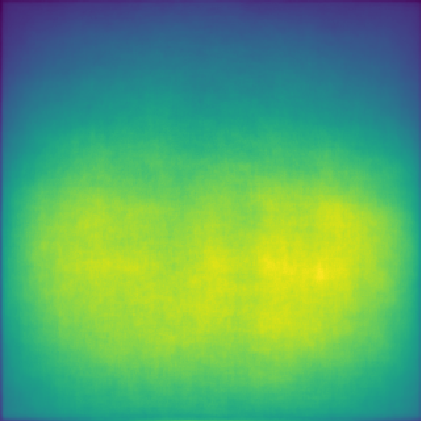}&
            \includegraphics[width=0.17\textwidth]{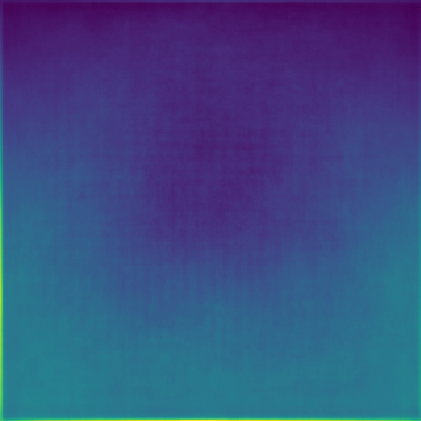}\\
             \multicolumn{2}{c} {\includegraphics[width=0.345\textwidth, height=0.29cm]{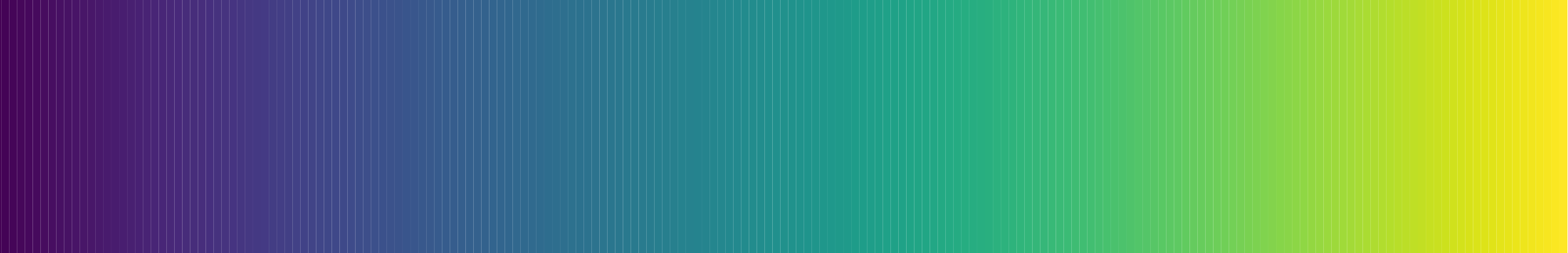}} \\
            
            Saliency~\cite{qin2019basnet} & Objectness \\
	      
	        \multicolumn{2}{c}{(b)} \\

			\hspace{0.08cm}
			
			\end{tabular}
		
		        \setlength\tabcolsep{5.8pt}
		        \def\arraystretch{1.25}
		    \begin{tabular}{ c| c | c}
 				\specialrule{1.2pt}{1pt}{1pt}
                
                * & $\mathcal{G}^{^{\mathcal{C}_m}}$ &  $\mathcal{G^{B}}$\\
                \specialrule{1.2pt}{1pt}{1pt}
                
                PicaNet & 53.7 & 64.7\\
                BASNet & 58.5 & 64.0 \\
                \hline
                Ours & \textbf{70.6} & \textbf{76.6} \\


				
 				\specialrule{1.2pt}{1pt}{1pt}
                 \multicolumn{3}{c}{(c)} \\
 			\end{tabular}
			}

			\caption{ (a) Visual comparison between our class agnostic objectness model and top performing saliency detectors. (b)  Saliency and objectness error distribution across the predictions on VOC12 training set. (c) Pseudo-label quality comparison in terms of mIoU on the VOC12 train set, between the objectness model and saliency methods.}
			\label{fig:no_voc}
			\end{center}
		\vspace{-0.6cm}
\end{figure}

\noindent\textbf{Improving Semantic Proposals: Objectness or Saliency Guidance? 
} \label{sec:sod}
It is common to utilize pixel-level \textit{saliency} information as additional guidance to be combined with the CAM proposals~\cite{oh2017exploiting,wei2018revisiting,zeng2019joint,yao2020saliency}. Specifically, DHSNet~\cite{liu2016dhsnet} and DSS~\cite{hou2016deeply} have been used in~\cite{hou2018self,chaudhry2017discovering,wei2018revisiting} to generate a saliency mask for each training sample. This guidance of saliency can deliver non-semantic pixel-level supervision for a better boundary segmentation. However, the saliency information used in the previous studies only focus on the \textit{most} salient object due to the problem of centre bias~\cite{bruce2015computational, accik2014real}. For instance, as shown in Fig.~\ref{fig:no_voc}(a), the masks generated by saliency models can only detect the objects near the centre of an image, the ``ship'' near the corner will be incorrectly labelled as background (top row). Furthermore, the region of the ``train'' can only be partially labelled because the back of the train is not salient. This problem introduces outliers (incorrectly labelled regions) when training a segmentation model.
In contrast, our proposed objectness model learns to recognize objects in all image locations, even if they are not salient or near the image boundary, see Fig.~\ref{fig:no_voc}(a). Figure~\ref{fig:no_voc}(b) further illustrates that the objectness network is equally likely to make errors in all image locations, while the saliency detection network is biased towards making erroneous predictions near the image border. To validate this claim quantitatively, we conduct experiments (see Table in Fig.~\ref{fig:no_voc}(c)) by replacing the objectness mask with a saliency mask for creating semantic pseudo-labels. We use two recent saliency detectors, PiCaNet~\cite{liu2018picanet} and BASNet~\cite{qin2019basnet} to generate the saliency mask for VOC12 training images. Combining saliency masks with $\mathcal{G}^{^{\mathcal{C}_m}}$ and $\mathcal{G^{B}}$ achieves performance which is significantly lower than the quality of our pseudo-labels generated using the objectness guidance. 

\vspace{-0.3cm}
\section{Discussion and Conclusion}
\blue{Existing saliency-based WSSS~\cite{papandreou2015weakly,huang2018weakly,lee2019ficklenet,zeng2019joint,jiang2019integral,fan2020employing,fan2020learning} and SSSS~\cite{lee2019ficklenet} methods utilize both saliency detectors (trained on the DUT-S~\cite{wang2017learning} or MSRA-B~\cite{WangDRFI2017} datasets which have pixel-level binary segmentation ground-truth for a large number of overlapping instances in the VOC12 dataset) and a portion of semantic segmentation GT, respectively. Following this line of work, we choose the objectness-based dataset to introduce a better proposal model which addresses the severe center bias issue of saliency detectors (see Fig.~\ref{fig:no_voc} (a, b)) for WSSS (e.g., saliency inherently ignores objects near the border). We compare with both WSSS and SSSS techniques since we do not fall neatly within either category of supervision (i.e., comparing against methods which use only CAM is unfair but we also do not use any semantic segmentation GT).}
Moreover, in contrast to the previous methods~\cite{khoreva2017simple,xiao2018transferable,ahn2018learning,ahn2019weakly}, our framework does not require multiple stages of label inference and training for pseudo-label generation, but instead operates in a single stage. 
Additionally, the objectness branch improves the performance of the segmentation network by propagating boundary and semantic information back through the network. We believe the objectness branch helps with semantics because it forces the model to treat objects more uniformly (since the objectness label is binary). This can guide the segmentation model to treat nearby pixels as the same semantic object class and promote more spatially uniform predictions, which is correct in many cases. 

In summary, we have presented a pseudo-label generation and joint learning strategy for the tasks of both WSSS and SSSS. We first introduced a novel technique to generate high quality pseudo-labels that combines class agnostic objectness priors with either image-level labels or bounding box annotations. Next, we proposed a model that jointly learns semantics and objectness to guide the network to encode more accurate boundary information and better semantic representations. We conducted an extensive set of experiments under different settings and supervision strategies to validate the effectiveness of the proposed methods. The ablation studies isolated the improvements due to the proposed objectness branch, and validated the efficacy of our ignoring strategy. Furthermore, the pseudo-label generation pipeline is simple, efficient, and can be used for large-scale data annotation. 

\clearpage
\newpage 


\newcommand{\hbAppendixPrefix}{S}
\renewcommand{\thefigure}{\hbAppendixPrefix\arabic{figure}}
\setcounter{figure}{0}
\renewcommand{\thetable}{\hbAppendixPrefix\arabic{table}}
\setcounter{table}{0}
\renewcommand{\theequation}{\hbAppendixPrefix\arabic{equation}}
\setcounter{equation}{0}
\renewcommand{\thesection}{\hbAppendixPrefix\arabic{section}}
\setcounter{section}{0}
\newcommand{\size}[2]{{\fontsize{#1}{0}\selectfont#2}}

\begin{center}
\vspace{10cm}
\textbf{\size{16}{\bluee{Simpler Does It: Generating Semantic Labels with Objectness Guidance}} }   
\vspace{0.5cm}
    
\textbf{\size{14}{\bluee{--Supplementary Materials--} }}      
\end{center}

\begin{figure} [h]
	\begin{center}
			\includegraphics[width=0.99\textwidth]{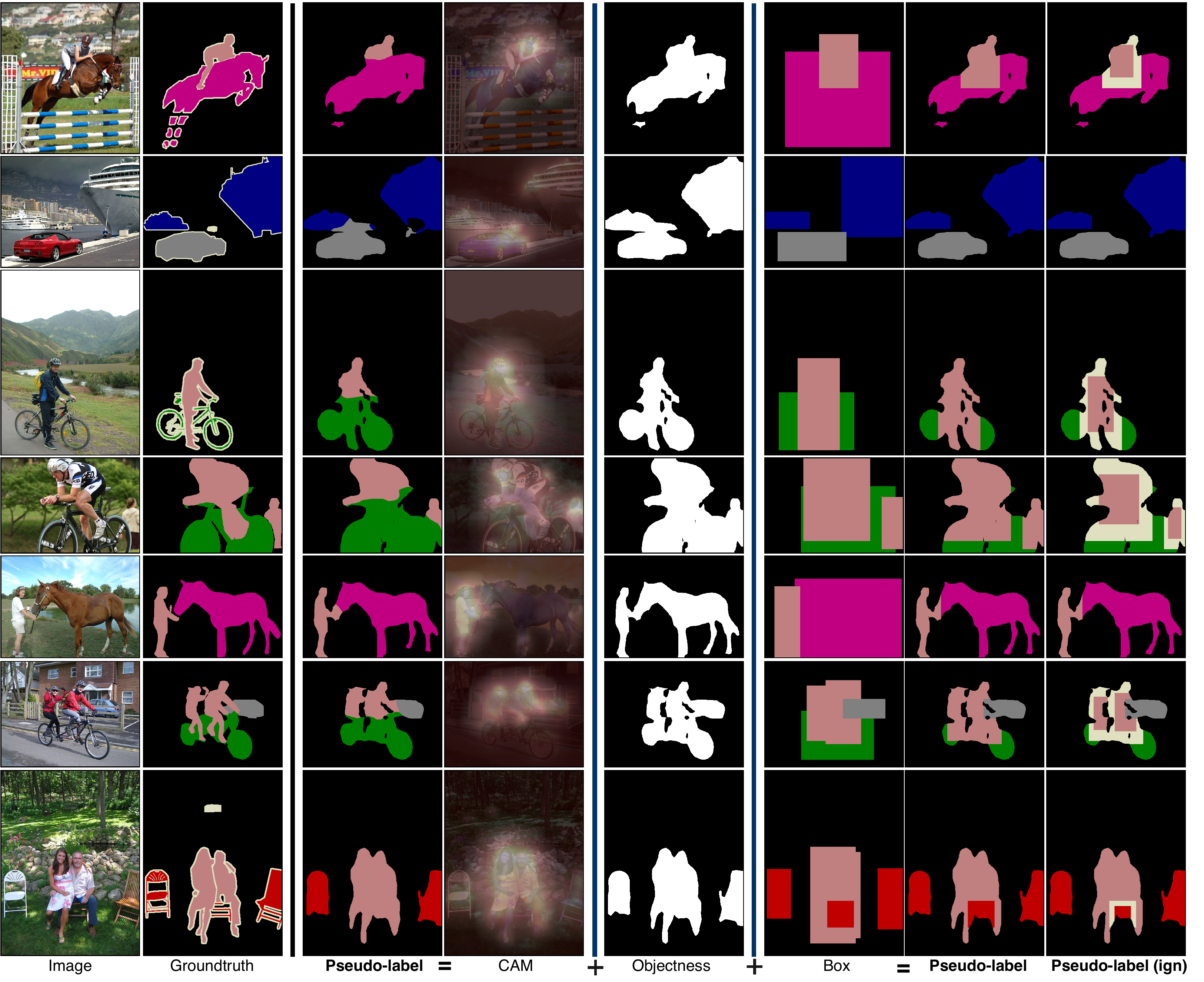}
		\vspace{-0.1cm}
		\caption{Additional examples of our pseudo-label generation process. We combine a class agnostic objectness prior with either a Class Activation Maps (CAMs)~\cite{zhou2016learning} \textit{or} bounding box proposal to generate a pseudo-label.}
		\label{fig:pseudo_ignore}
		\vspace{-0.5cm}
	\end{center}
\end{figure}

\section{Examples of Generated Pseudo-labels on VOC 2012}\label{sec:exp}
Figure~\ref{fig:pseudo_ignore} shows additional examples of the generated pseudo-labels by combining the class agnostic objectness priors with either CAM~\cite{zhou2016learning} or bounding box proposals. Our pseudo-label generation technique successfully extracts boundary information from the objectness prior and class information from the CAM or bounding box proposals, resulting in high-quality pseudo-labels with fine-grained details about the object's shape. For the ignore strategy, we assign values of 255 to the outer regions of a bounding box if it overlaps (above a certain threshold) with the largest bounding box in an image. These overlapped semantic regions have a high degree of uncertainty due to the inherent structure of bounding boxes and ignoring these regions during training results in better predictions (see Fig.~\ref{fig:voc12val}). 

\section{Details of SONet Architecture}\label{sec:exp}

\blue{We discussed the SONet architecture in Sec. 2.2 of the main manuscript. The details of the objectness module in SONet architecture are shown in Table~\ref{tab:vgg5}. The input to the objectness module is the segmentation map, $\mathcal{S}\in \mathbb{R}^{b\times 21\times16\times16}$ which is generated by the DeepLabv3-ResNet101 network. The objectness module consists of five convolution layers where first four layers gradually increase the depth (i.e., channel) of the feature map.  The last convolution layer predicts the desired objectness map, $\mathcal{O}\in \mathbb{R}^{b\times 2\times16\times16}$. Note that we apply batch normalization and ReLU layers after each convolution layers except the last one which predicts the objectness map. The newly introduced convolution layers are trained from scratch.}
\begin{table} [t]

\def\arraystretch{1.25}
\centering

\resizebox{0.59\textwidth}{!}{
	\begin{tabular}{c }
	\specialrule{1.2pt}{1pt}{1pt}
	 
	 \textbf{Input:} Segmentation Map $\mathcal{S}\in \mathbb{R}^{b\times 21\times16\times16}$ \\
	 \midrule
	 
	 Conv2d ($3\times3$), Batch Norm, ReLU $\rightarrow$ $\mathbb{R}^{b\times 32\times16\times16}$ \\
	  \midrule
	  
	  Conv2d ($3\times3$), Batch Norm, ReLU $\rightarrow$ $\mathbb{R}^{b\times 64\times16\times16}$ \\
	   \midrule
	   
	    Conv2d ($3\times3$), Batch Norm, ReLU $\rightarrow$ $\mathbb{R}^{b\times 128\times16\times16}$ \\
	   \midrule
	   
	    Conv2d ($3\times3$), Batch Norm, ReLU $\rightarrow$ $\mathbb{R}^{b\times 256\times16\times16}$ \\
	   \midrule 
	   
	    Conv2d ($1\times1$), Batch Norm, ReLU $\rightarrow$ $\mathbb{R}^{b\times 2\times16\times16}$ \\
	    
	    \midrule 
	   
	    \textbf{Output:} Objectness Map $\mathcal{O}\in \mathbb{R}^{b\times 2\times16\times16}$ \\

	\specialrule{1.2pt}{1pt}{1pt}
	
	

		\end{tabular}
	
	}
	\caption{Configuration of the Objectness Module in SONet.}
    \label{tab:vgg5}
\end{table}

\section{Supplementary Experiments}\label{sec:exp}
In this section, we first provide implementation details of our proposed SONet (Sec.~\ref{sec:implement}) and a description of the OpenV5 dataset (Sec.~\ref{sec:openv5}). Then, we provide anonymous links to the PASCAL VOC 2012 test set results and additional qualitative examples predicted by SONet with different levels of supervision (Sec.~\ref{sec:img_seg}). Further we  conduct experiments on video object segmentation (Sec.~\ref{sec:video}). We also show the generality of our proposed pseudo-label generation technique on the Berkeley DeepDrive dataset~\cite{yu2018bdd100k} (Sec.~\ref{sec:deepdrive}). Finally, we report a series of ablation studies (Sec.~\ref{sec:SONet_abl}).

\subsection{Implementation Details}\label{sec:implement}
We implement our method using the PyTorch~\cite{paszke2019pytorch} framework trained end-to-end on two NVIDIA GeForce
GTX 1080 Ti GPUs. We use the SGD optimizer to train our network. We train all the variants of our SONet for 40 epochs with an initial learning rate of 2e-3. We use a random crop of 513$\times$513 and 321$\times$321 during training for SONet and SONet$^*$, respectively. Similarly, we use a output stride of 16 and 8 during training for SONet and SONet$^*$, respectively. During inference, we use a crop of 513$\times$513 and rescale to the original size using simple bilinear interpolation before calculating the mIoU. Following the current practice~\cite{chen2018deeplab,zhao2017pyramid,noh15_iccv}, to report test set results on PASCAL VOC 2012, we first train on the augmented training set followed by fine-tuning on the original trainval set with the generated pseudo-labels.
\begin{figure*}
	\begin{center}
			\includegraphics[width=0.9\textwidth]{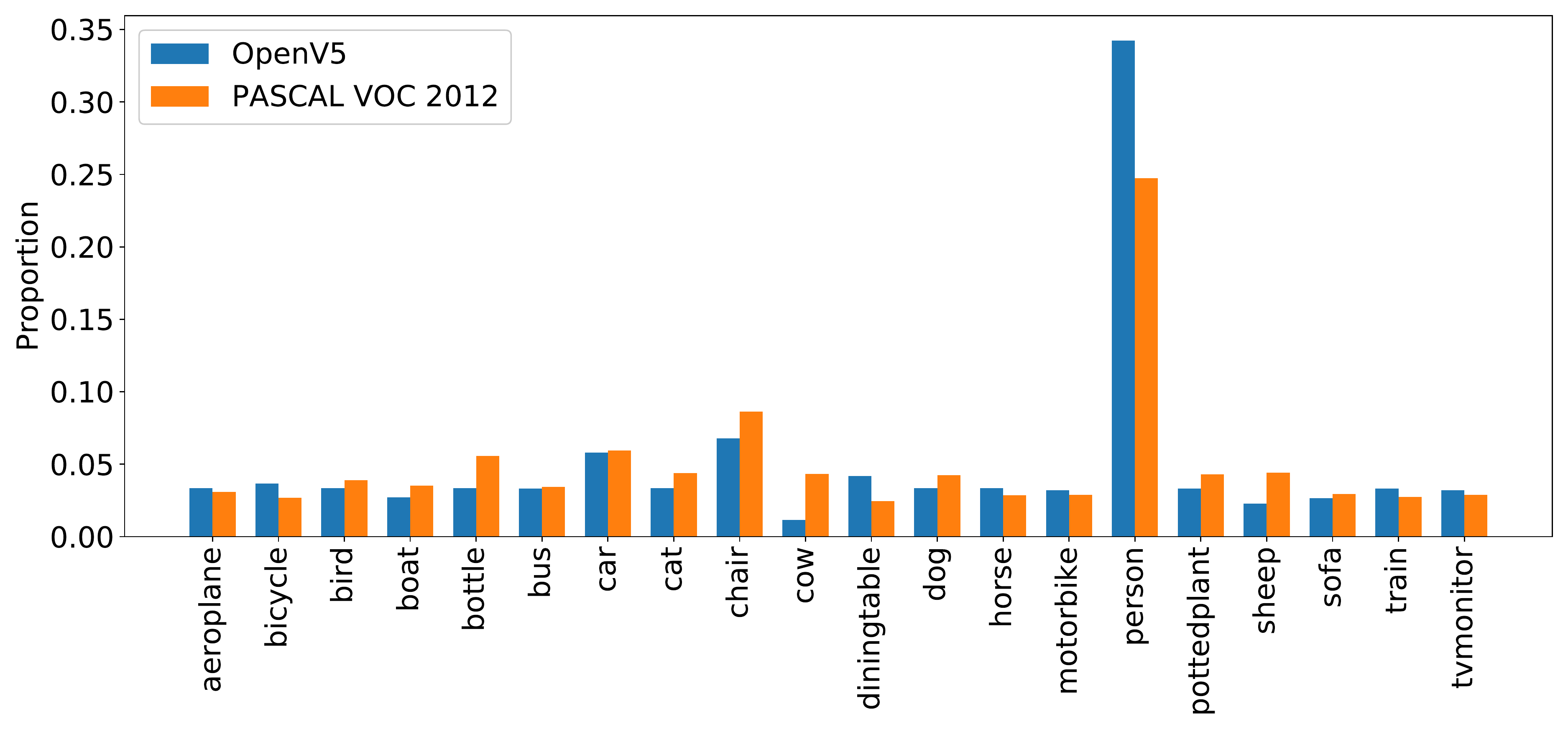}
		\vspace{-0.2cm}
		\caption{The distribution of semantic object categories in the OpenV5 subset and PASCAL VOC 2012 training sets.}
		\label{fig:openv5}
	\end{center}
\end{figure*}
\subsection{OpenV5 Dataset}\label{sec:openv5}
\begin{figure*}
	\begin{center}
		\includegraphics[width=1.0\textwidth]{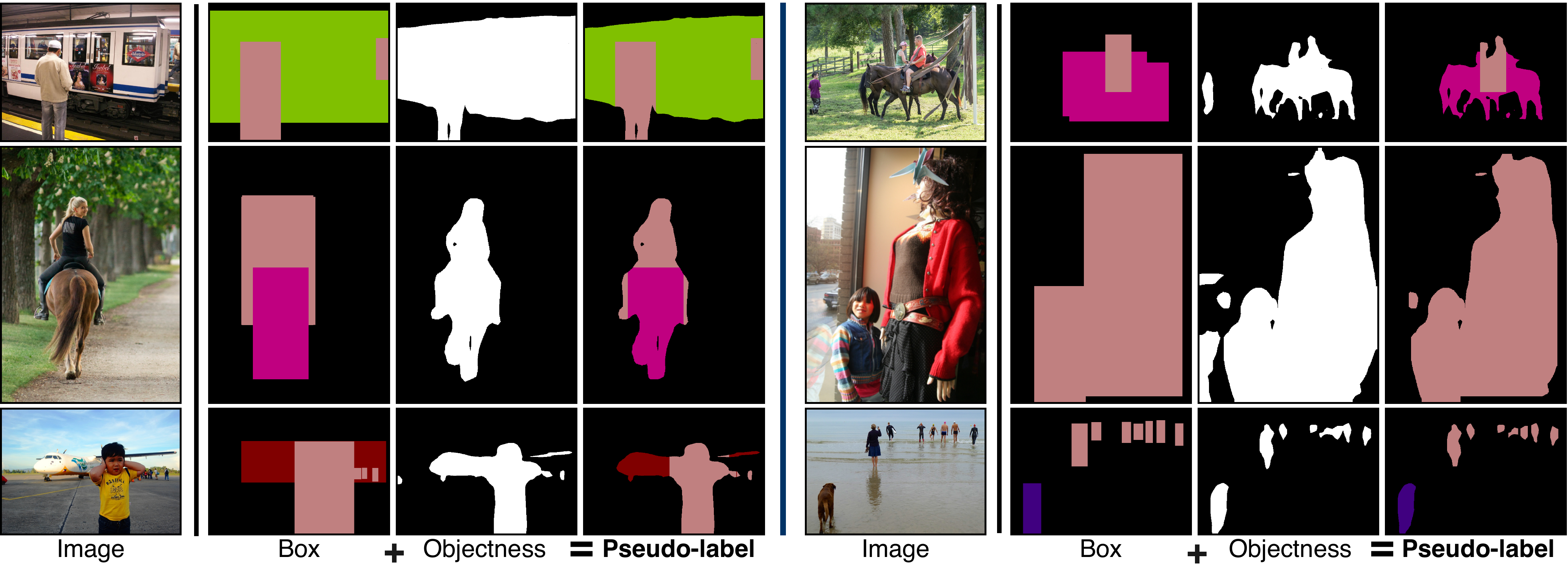}
		\vspace{-0.1cm}
		\caption{OpenV5 pseudo-labels. Examples of generated pseudo-labels with our proposed approach on the OpenV5 images.} 
		
		\label{fig:pseudo_openv5}
	\end{center}
\end{figure*}

We have shown experimental results using the OpenV5 dataset for the task of semantic segmentation in Table 1(c) of the main paper. We compare against the state-of-the-art by using the standard protocol (training on PASCAL VOC 2012 augmented train set and evaluate on PASCAL VOC 2012 val/test set). As mentioned in Sec. 3.1 of the main manuscript, we use a subset of the OpenV5 dataset, where each semantic category is contained in a large number of images, consisting of 42,621 total images and 20 semantic categories. Figure~\ref{fig:openv5} shows the comparison of the object instance distribution of the OpenV5 subset and PASCAL VOC 2012 dataset. It is evident from the table that there are a considerable number of instances for each semantic category and \textit{person} is by far the most dominant category as expected since it co-occurs with most of other categories. Figure~\ref{fig:pseudo_openv5} shows examples of the generated pseudo-labels for OpenV5, by combining the class agnostic objectness priors with bounding box proposals.

\begin{figure*}
	\begin{center}
		\includegraphics[width=0.98\textwidth]{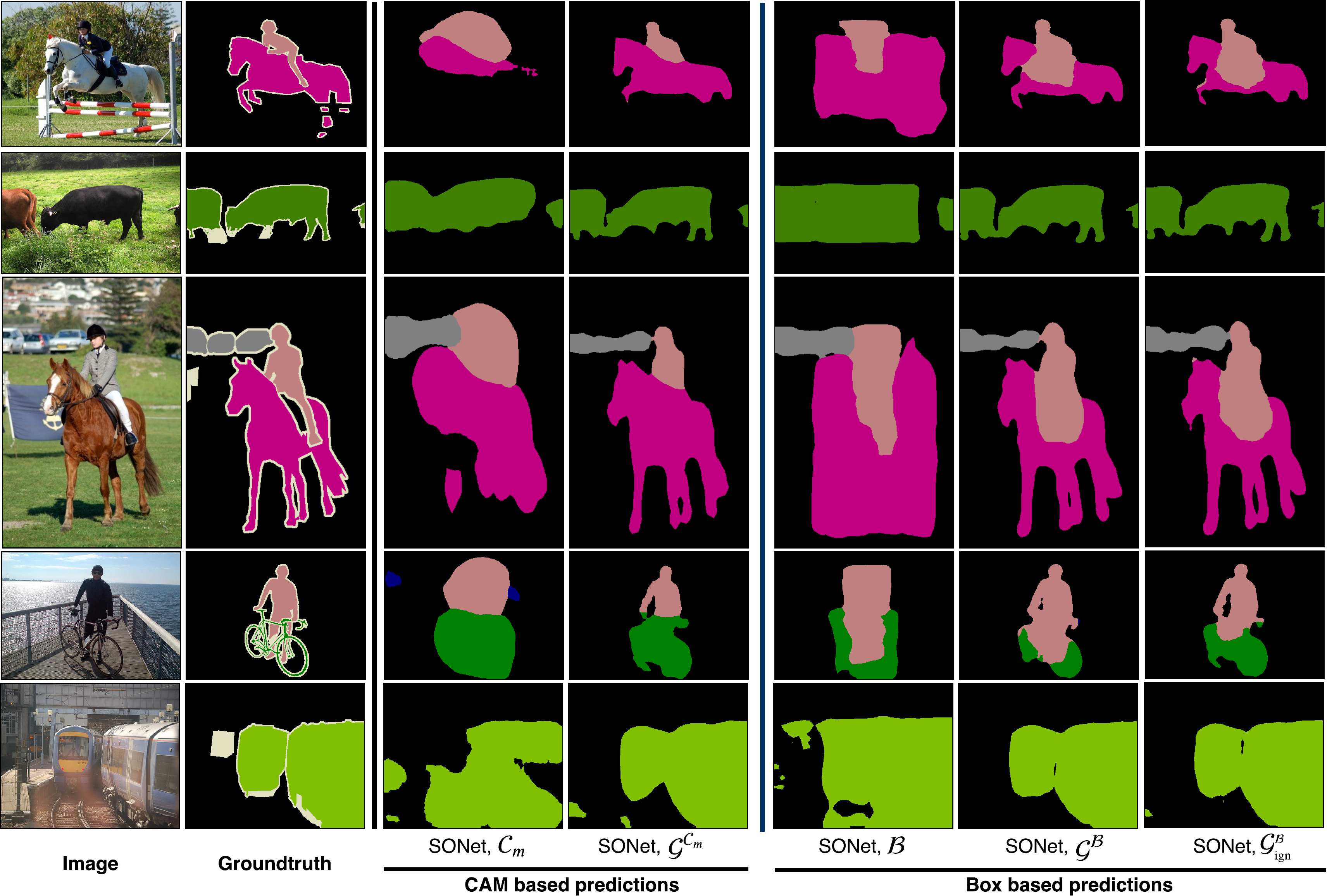}
		\caption{\textbf{Examples of predicted segmentation maps} with our proposed approach on PASCAL VOC 2012 validation images.} 
		
		\label{fig:voc12val}
	\end{center}
\end{figure*}

				
			    
			    
			    
				   
				   
				   


			


\subsection{PASCAL VOC 2012 Test Set Results}\label{sec:img_seg}
We illustrate additional visual examples predicted by SONet with different levels (CAMs and box-driven) of supervision on PASCAL VOC 2012 validation images in Fig.~\ref{fig:voc12val}. The segmentation mask generated by SONet produces more accurate results when trained with CAMs or box-driven pseudo-labels than SONet trained solely with CAMs or bounding box annotations.
\subsection{Video Object Segmentation Results}\label{sec:video}
We also experiment on the YouTube-Object (YTO) dataset~\cite{prest2012learning} to show the effectiveness of our method in segmenting objects from videos by simply evaluating the results produced by SONet. 
Following prior works~\cite{tang2013discriminative,papazoglou2013fast,lee2019frame}, we use the groundtruth segmentation masks provided by~\cite{jain2014supervoxel} to evaluate the performance of SONet and also compare our method with recent video segmentation methods with weak supervision in Table~\ref{tab:youtube}. Note that all the baseline methods are explicitly trained on video datasets and use temporal cues, while our method is trained on static images without temporal information. Our SONet method outperforms the existing methods which use different levels of supervision. This may be because objectness-driven pseudo-labels provide more fine-grained localization with sharper object boundaries than coarse bounding boxes. Samples of the predicted masks for the YTO dataset are shown in Fig.~\ref{fig:youtube}.  
\begin{table} [t]
\begin{center}
        \def\arraystretch{1.3}
		\resizebox{0.64\textwidth}{!}{
		\begin{tabular}{l|  c cccccc |c}

				&\rotatebox{90}{SOSD~\cite{zhang2015semantic}} 
				&\rotatebox{90}{OVS~\cite{drayer2016object}} 
				&\rotatebox{90}{DPM~\cite{zhang2017semantic}}  
			    &\rotatebox{90}{BBF~\cite{saleh2017bringing}}  
				&\rotatebox{90}{Crawl~\cite{hong2017weakly}}  
				&\rotatebox{90}{SROW~\cite{yang2018segmentation}}  
				&\rotatebox{90}{AAR~\cite{lee2019frame}}  
				&\rotatebox{90}{\textbf{SONet} } \\
				
				\specialrule{1.2pt}{1pt}{1pt}
				
				Temporal &  $\checkmark$& $\checkmark$& $\checkmark$& $\checkmark$& $\checkmark$& $\checkmark$ & $\checkmark$ & X  \\
				
				Sup.& $\mathcal{B}$ & $\mathcal{B}$ & $\mathcal{B}$ & $\mathcal{I}$ & $\mathcal{I}$ & $\mathcal{I}$ & $\mathcal{I}$ & $\mathcal{I}$  \\
				mIoU& 54.1 & 56.2 & 61.7 & 53.3 & 58.6 & 61.9 & 62.1 & \textbf{64.3}\\

				\specialrule{1.2pt}{1pt}{1pt}
	   \end{tabular} 
	   
	   }
	   \caption{Quantitative comparison of recent video object segmentation methods with various methods of supervision on the YouTube-Object dataset. $\mathcal{B}$: bounding box, $\mathcal{I}$: image-level supervision. Note that the baseline numbers are taken from~\cite{lee2019frame} for fair comparison.}
	\label{tab:youtube}
	\vspace{-0.3cm}
\end{center}
\end{table}
\begin{figure}
\begin{center}
		\resizebox{0.5\textwidth}{!}{
		\def\arraystretch{1.3}
        \setlength\tabcolsep{0.4pt}
		\begin{tabular}{*{3}{c}}		
				\fcolorbox{black}{black}{\includegraphics[width=0.3\textwidth,height=3.2cm]{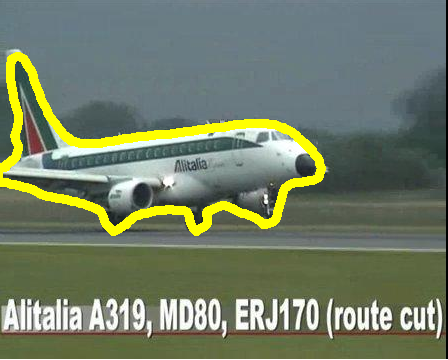}}&
				\fcolorbox{black}{black}{\includegraphics[width=0.3\textwidth,height=3.2cm]{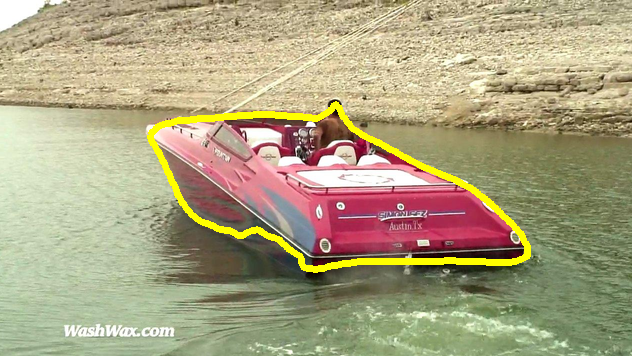}}&
				\fcolorbox{black}{black}{\includegraphics[width=0.3\textwidth,height=3.2cm]{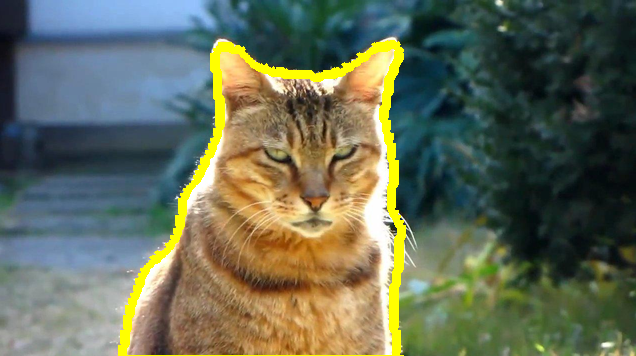}}\\
				\fcolorbox{black}{black}{\includegraphics[width=0.3\textwidth,height=3.2cm]{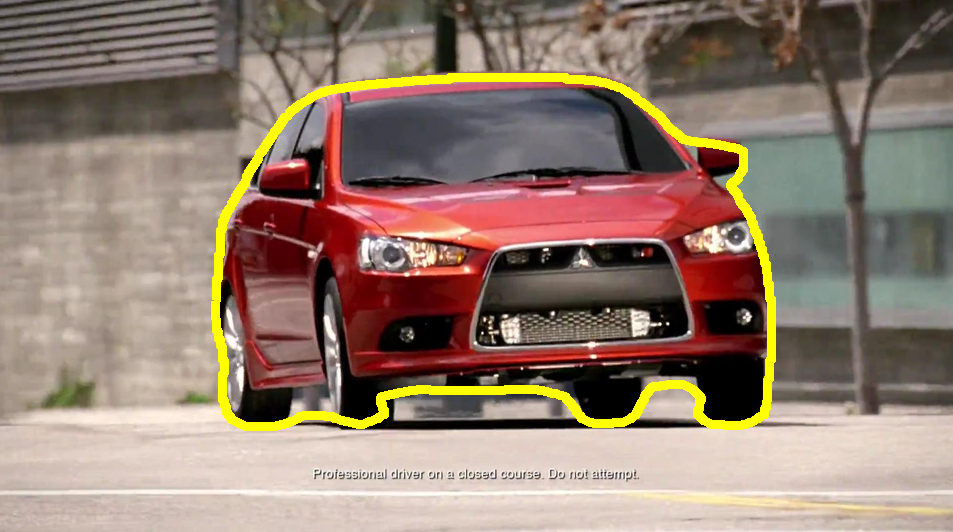}}&
				\fcolorbox{black}{black}{\includegraphics[width=0.3\textwidth,height=3.2cm]{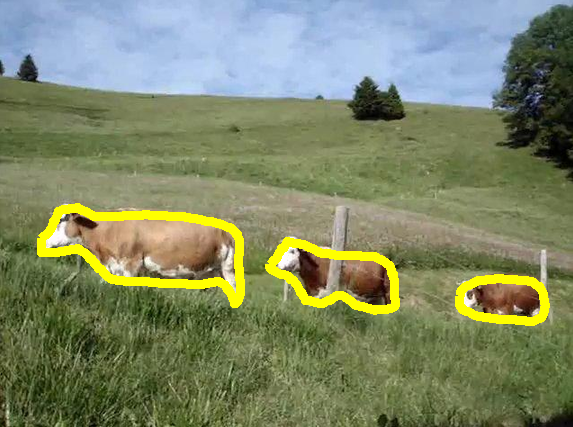}}&
				\fcolorbox{black}{black}{\includegraphics[width=0.3\textwidth,height=3.2cm]{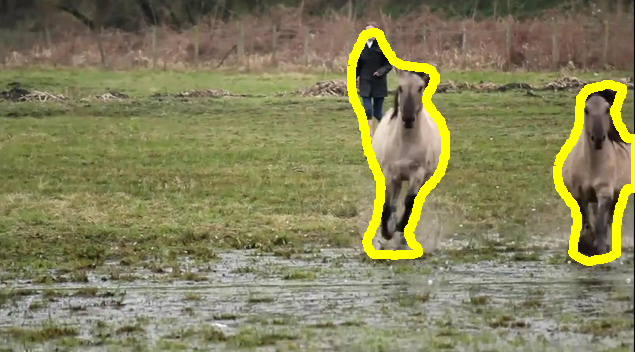}}\\

	 \end{tabular}}
		\caption{Predicted masks for frames of the YouTube-Object dataset. The outline of segmented regions are shown as yellow curves. Note that the object boundaries of SONet's predictions are detailed and smooth.}
	\label{fig:youtube}
	\end{center}
	\vspace{-0.4cm}
\end{figure}

\subsection{Generalization to Different Domains: Berkeley DeepDrive}\label{sec:deepdrive} 
\begin{figure*}
	\begin{center}
		\includegraphics[width=1.0\textwidth]{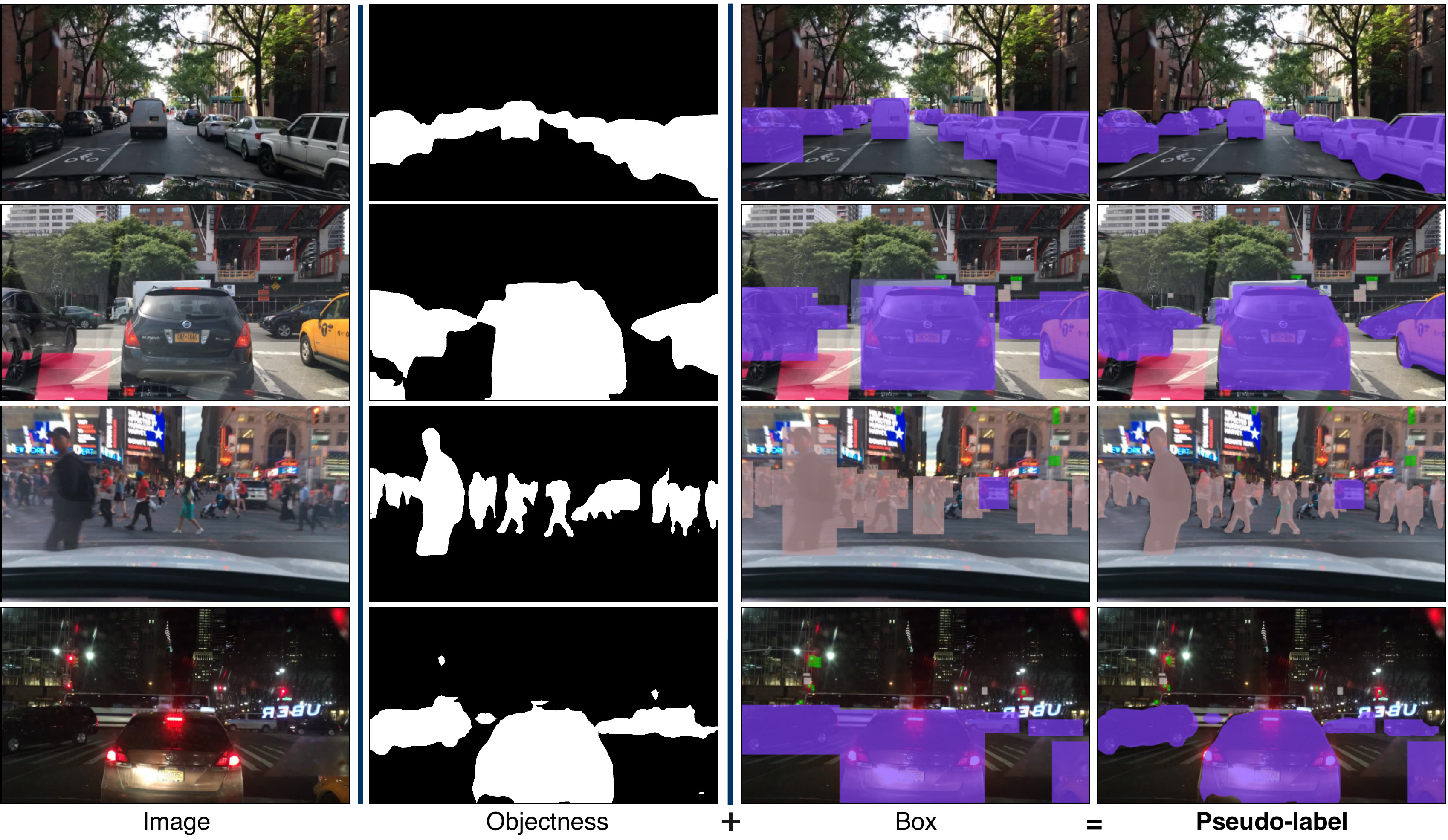}
		\vspace{-0.4cm}
		\caption{\textbf{Berkeley DeepDrive pseudo-labels}. Examples of generated pseudo-labels with our proposed approach on the Berkeley DeepDrive video frames.} 
		
		\label{fig:deepdrive}
	\end{center}
	\vspace{-0.4cm}
\end{figure*}

We further apply our bounding box-driven pseudo-label generation technique on a recent driving dataset, Berkeley DeepDrive~\cite{yu2018bdd100k}, to validate whether our procedure can generalize well on a dataset from a different domain. The Berkeley DeepDrive dataset~\cite{yu2018bdd100k} is composed of images of diverse road scenes (with motion blur) taken from various locations throughout the USA. We generate pseudo-labels for 100k frames which have bounding box annotation available for the 10 different categories: bus, light, sign, person, bike, truck, motor, car, train, and rider. Figure~\ref{fig:deepdrive} presents examples of generated pseudo-labels of DeepDrive video frames. It is clear that our class agnostic objectness model can generate masks with sharp boundaries in complex driving scenarios, resulting in high-quality pseudo-labels. Since the DeepDrive dataset does not provide pixel-wise annotation for these 100k frames we can not evaluate the quality of generated pseudo-labels in terms of mIoU.  
\subsection{Ablation Studies}\label{sec:SONet_abl} 
We conduct further ablation studies to analyze our design and the effectiveness of the objectness branch (Sec.~\ref{sec:designobj} \& Sec.~\ref{sec:objqual}).

\subsubsection{Design Choices of Objectness Branch.}\label{sec:designobj}
We vary the design of the objectness branch, $\phi_O$, of SONet and compare the architectures against each other. The results are reported in Table~\ref{tab:abl_branch2}. We evaluate three different variants: (v1) a single 1$\times$1 convolutional layer which predicts the objectness and takes as input the final feature representation (\texttt{res5C}), (v2) a single 1$\times$1 convolution layer which takes as input the semantic prediction ($\mathcal{S}$), and (v3) a smaller network is applied (as discussed in Sec. 3.2 of the main manuscript) but takes as input the features from \texttt{res5C}.

\subsubsection{Effectiveness of Objectness Branch in SONet}\label{sec:objqual}
\begin{figure} [t]
	\begin{center}
		\setlength\tabcolsep{0.5pt}
		\def\arraystretch{0.5}
		\resizebox{0.98\textwidth}{!}{
			\begin{tabular}{*{5}{c}}

	      	\includegraphics[width=0.18\textwidth]{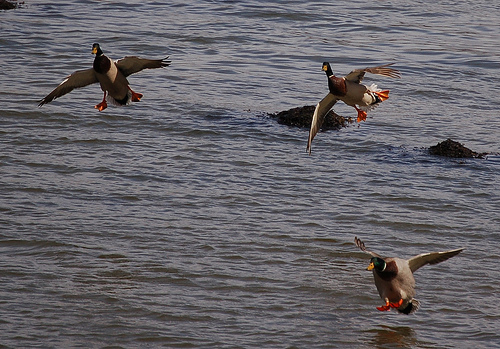}&
	       \includegraphics[width=0.18\textwidth]{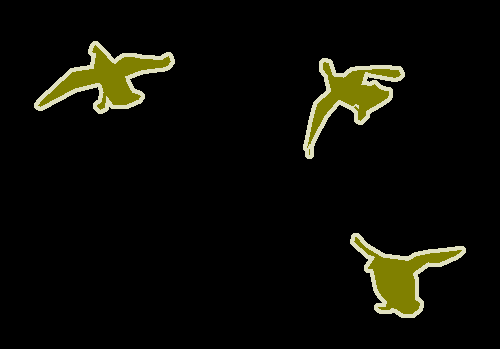}&	
	       \includegraphics[width=0.18\textwidth]{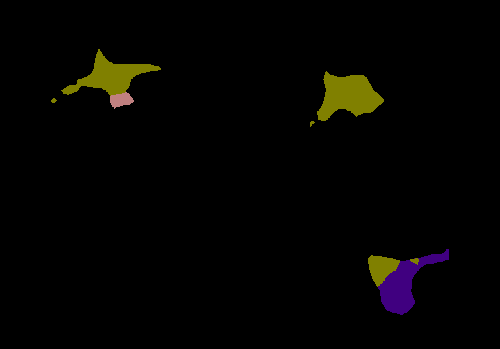}&
	       \includegraphics[width=0.18\textwidth]{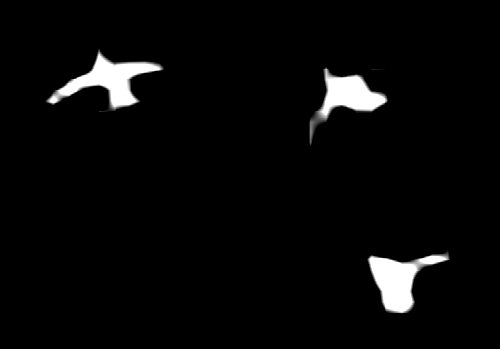}&
	      \includegraphics[width=0.18\textwidth]{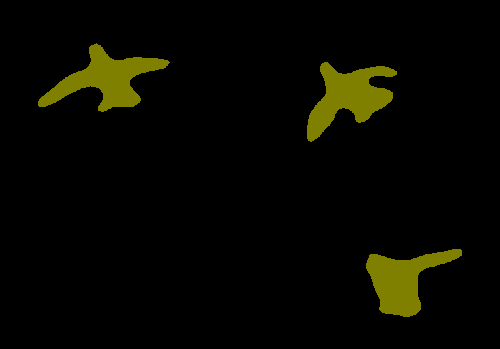}\\
	      
	      	\includegraphics[width=0.18\textwidth]{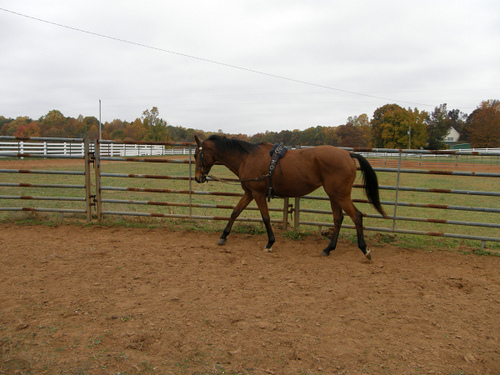}&
	       \includegraphics[width=0.18\textwidth]{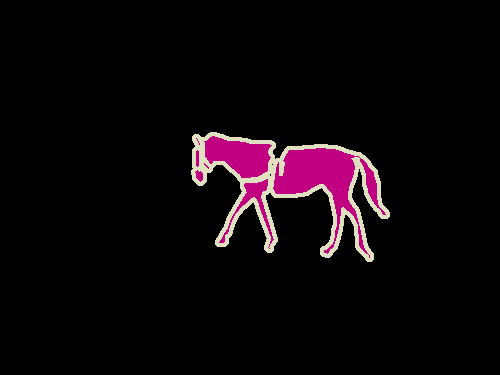}&	
	       \includegraphics[width=0.18\textwidth]{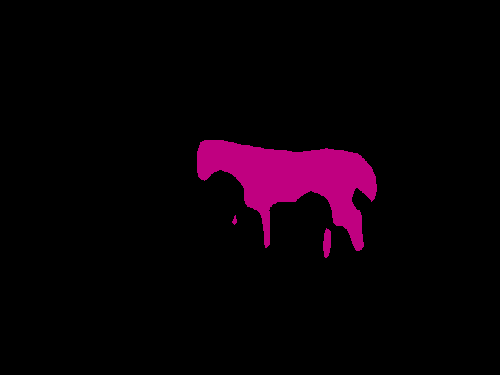}&
	       \includegraphics[width=0.18\textwidth]{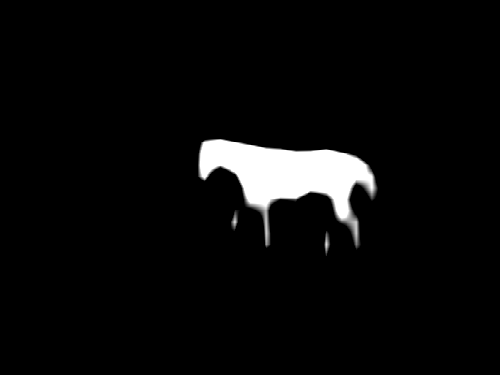}&
	      \includegraphics[width=0.18\textwidth]{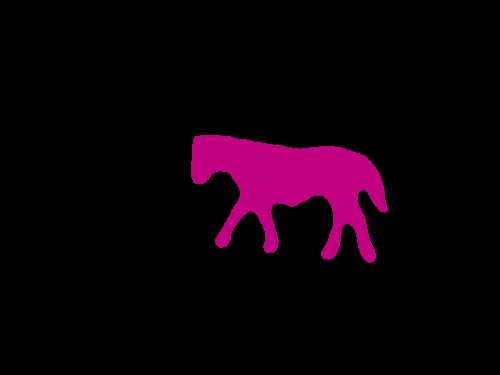}\\

	      \includegraphics[width=0.18\textwidth]{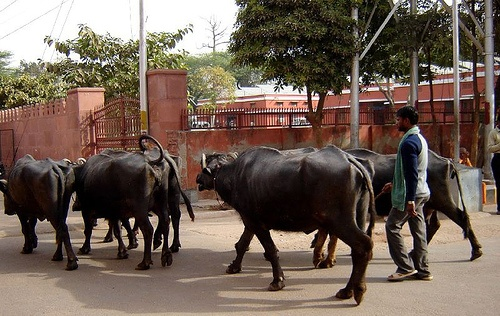}&
	       \includegraphics[width=0.18\textwidth]{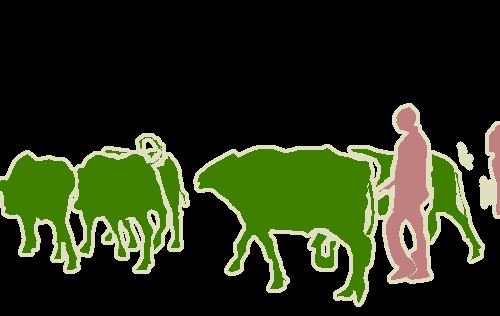}&		
	       \includegraphics[width=0.18\textwidth]{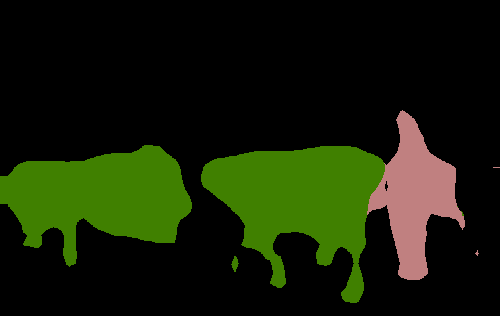}&
	       \includegraphics[width=0.18\textwidth]{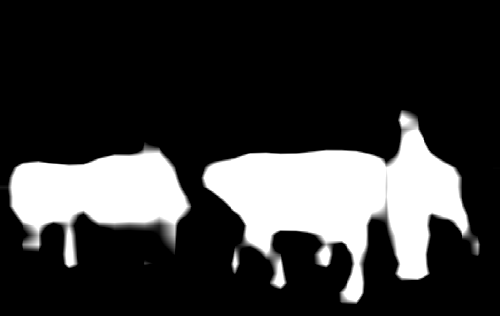}&
	      \includegraphics[width=0.18\textwidth]{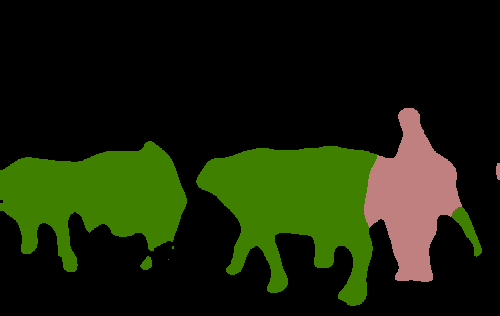}\\
	      
	      	\includegraphics[width=0.18\textwidth]{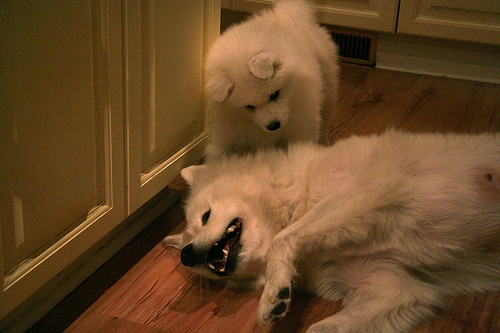}&
	       \includegraphics[width=0.18\textwidth]{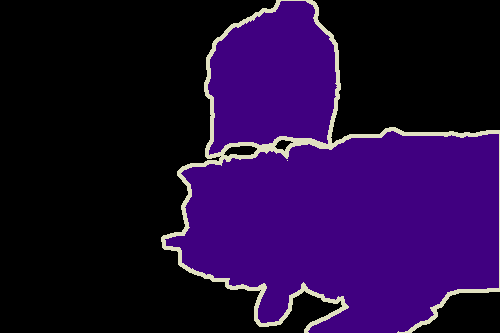}&		
	       \includegraphics[width=0.18\textwidth]{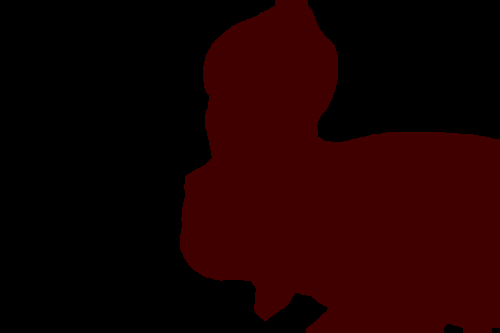}&
	       \includegraphics[width=0.18\textwidth]{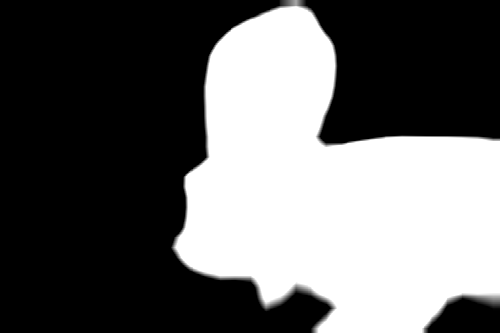}&
	      \includegraphics[width=0.18\textwidth]{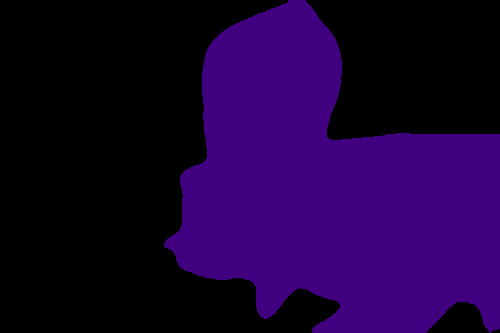}\\

	      \includegraphics[width=0.18\textwidth]{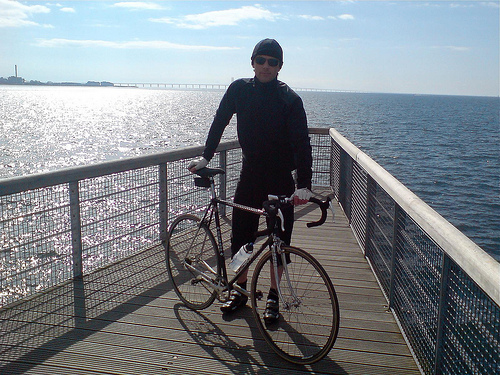}&
	       \includegraphics[width=0.18\textwidth]{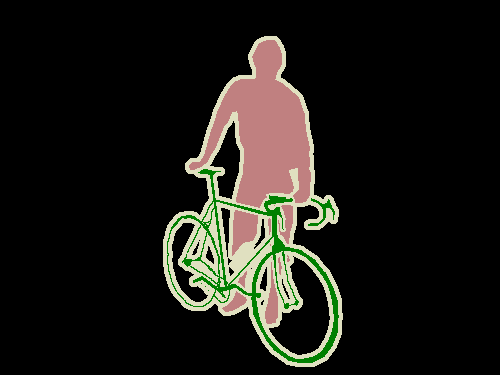}&		
	       \includegraphics[width=0.18\textwidth]{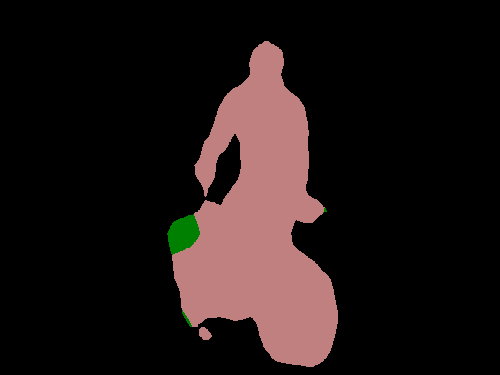}&
	       \includegraphics[width=0.18\textwidth]{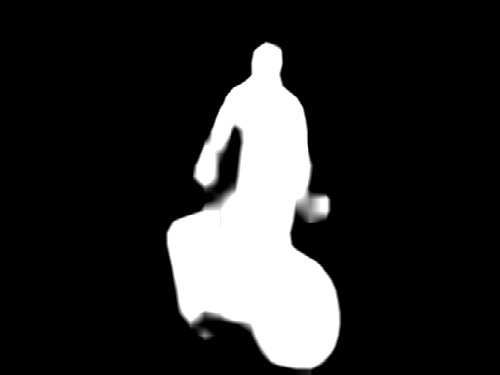}&
	      \includegraphics[width=0.18\textwidth]{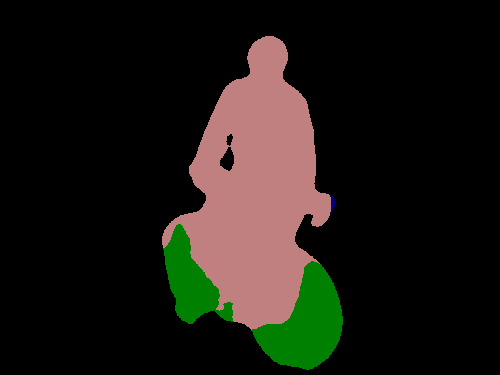}\\

		  Image & GT & DeepLabv3~\cite{chen2018deeplab} & Objectness & SONet\\	
			
			\end{tabular}

			}
			
		\end{center}
		\caption{\textbf{Visualization of the effect of the objectness branch} on the segmentation results. Note that the difference between the baseline DeepLabv3~\cite{chen2018deeplab} and SONet is the objectness branch. Images are taken from the PASCAL VOC 2012 validation set.}
		\label{valobj2}
\end{figure}

We provide additional qualitative examples in Fig.~\ref{valobj2} to show the objectness branch's effect
on SONet's semantic segmentation predictions. Note that SONet without the objectness
branch is equivalent to DeepLabv3~\cite{chen2018deeplab}. As can be seen from the examples, the objectness branch can guide the segmentation network to produce more accurate and smooth predictions.  
\begin{table} [t]
	\begin{center}
		\resizebox{0.85\textwidth}{!}{
			\begin{tabular}{c | c |c| c |c }
				
				\specialrule{1.2pt}{1pt}{1pt}
                \rowcolor{maroon!7}
                
                \multicolumn{1}{c|}{Name} & Sup.& Architecture ($\phi_O$) & Input & mIoU \\
                \specialrule{1.2pt}{1pt}{1pt}

                \textbf{SONet} & $\mathcal{G^B}$&smaller network discussed in Sec. 3.3 & \texttt{semantic} ($\mathcal{S}$)&\textbf{73.8} \\
                \hline
                v1 & $\mathcal{G^B}$&single 1$\times$1 convolution layer & \texttt{res5C} &72.2 \\
                 v2 &$\mathcal{G^B}$& single 1$\times$1 convolution layer & \texttt{semantic} ($\mathcal{S}$)& 73.5 \\
                 v3 &$\mathcal{G^B}$& smaller network discussed in Sec. 3.3 & \texttt{res5C} & 73.6 \\

				\specialrule{1.2pt}{1pt}{1pt} 
				
			\end{tabular}
				
			

				
				

			}
			\caption{Comparison of objectness branch variants for SONet on the PASCAL VOC 2012 validation set.}
			\label{tab:abl_branch2}
		\end{center}
\end{table}

\subsubsection{Transferring Semantic Knowledge from Source to Target Dataset.} As an additional baseline, we directly transfer the semantic information from COCOStuff to the VOC12 dataset. Towards this goal, we first train DeepLabv3~\cite{chen2017rethinking} on COCOStuff to output semantic segmentation (i.e., multi-class) masks instead of objectness masks (i.e., binary). Note, similar to the objectness training, we only consider the \textit{things} classes and use the pretrained model to generate pseudo-label (quality: 50.8\% mIoU) for the VOC12 train set. Then, we train DeepLabv3 using the generated pseudo-labels, resulting in 53.4\% mIoU on VOC12 val set.

\clearpage
\newpage 

\bibliography{weakly,paper,semi}

\end{document}